\documentclass[a4paper,fleqn]{cas-sc}

\usepackage{apacite}
\usepackage[authoryear]{natbib}

\usepackage{url}
\usepackage{multirow}
\usepackage{listings}
\usepackage{algorithm}
\usepackage{algpseudocode}
\usepackage{booktabs}
\usepackage[most]{tcolorbox}
\usepackage{subcaption}
\usepackage{amssymb}
\usepackage{amsmath}
\usepackage{graphicx} 

\usepackage{algorithm}
\usepackage{algpseudocode}
\usepackage{rotating} 

\usepackage{xcolor}

\usepackage{hyperref}    
\usepackage[nameinlink,noabbrev]{cleveref}  

\definecolor{bestcolor}{rgb}{0.55,0.00,0.10}    
\definecolor{secondcolor}{rgb}{0.10,0.25,0.55}  

\usepackage{graphicx}
\usepackage{subcaption}

\usepackage{xspace}

\newcommand{\TopM}{\textsc{TopM}\xspace}
\newcommand{\TopK}{\textsc{TopK}\xspace}

\newtcolorbox{grayquote}{
    colback=gray!10,
    colframe=white,
    boxrule=0pt,
    left=5pt,
    right=5pt,
    top=5pt,
    bottom=5pt
}

\def\tsc#1{\csdef{#1}{\textsc{\lowercase{#1}}\xspace}}
\tsc{WGM}
\tsc{QE}
\tsc{EP}
\tsc{PMS}
\tsc{BEC}
\tsc{DE}

\begin{document}
\let\WriteBookmarks\relax
\def\floatpagepagefraction{1}
\def\textpagefraction{.001}

\shorttitle{Less is More for RAG}

\shortauthors{Song et~al.}

\title [mode = title]{Less is More for RAG: Information Gain Pruning for Generator-Aligned Reranking and Evidence Selection} 

\author[dut]{Zhipeng~Song}
[orcid=0009-0009-6249-1988]
\ead{songzhipeng@mail.dlut.edu.cn}

\author[dlou]{Yizhi~Zhou}
[orcid=0000-0002-6761-5953]
\ead{zhouyizhi@dlou.edu.cn}

\author[ldu]{Xiangyu~Kong}
[orcid=0000-0003-1940-8674]
\ead{xiangyukong@liaodongu.edu.cn}

\author[dut,qhu]{Jiulong~Jiao}
[orcid=0009-0001-9852-7999]
\ead{jiaojiulong@mail.dlut.edu.cn}

\author[ltu]{Xinrui~Bao}
[orcid=0009-0000-7916-2122]
\ead{4724200573@stu.lntu.edu.cn}

\author[dut]{Xu~You}
[orcid=0009-0009-4583-7537]
\ead{m13050778592@mail.dlut.edu.cn}


\author[dmu]{Xueqing~Shi}
[orcid=0009-0008-3754-6396]
\ead{shixq@dmu.edu.cn}

\author[tencentdl]{Yuhang~Zhou}
\ead{ginozhou@tencent.com}

\author[dut]{Heng~Qi}
[orcid=0000-0002-8770-3934]
\cormark[1]
\ead{hengqi@dlut.edu.cn}

\cortext[cor1]{Corresponding author.}






\affiliation[dut]{organization={School of Computer Science and Technology, Dalian~University~of~Technology},
    addressline={No.2 Linggong Road, Ganjingzi District},
    city={Dalian},
    postcode={116024}, 
    country={China}}

\affiliation[dmu]{organization={College of Health-Preservation and Wellness, Dalian Medical University},
    addressline={No. 9 West Section of Lvshun South Road, Lvshunkou District},
    city={Dalian},
    postcode={116044},
    country={China}}

\affiliation[dlou]{organization={School of Information Engineering, Dalian Ocean University},
    addressline={No. 2-52, Heishijiao Street, Shahekou District},
    city={Dalian},
    postcode={116023}, 
    country={China}}

\affiliation[ldu]{organization={School of Information Engineering, Liaodong~University},
    addressline={No.116 Linjiang Back Street, Zhenan District},
    city={Dandong},
    postcode={118001}, 
    country={China}}

\affiliation[qhu]{
organization={Information Technology Center, Qinghai~University},
    addressline={251 Ningda Road, Chengbei District},
    city={Xining},
    postcode={810016}, 
    country={China}}

\affiliation[ltu]{
organization={School of Electronic and Information Engineering, Liaoning~Technical~University},
    addressline={188 Longwan South Street, Sijiatun District},
    city={Huludao},
    postcode={125105}, 
    country={China}}


\affiliation[tencentdl]{organization={Tencent (Dalian Northern Interactive Entertainment Technology Co., Ltd.)},
    addressline={21/F, Tencent Building, No. 26 Jingxian St, Ganjingzi District},
    city={Dalian},
    postcode={116085}, 
    country={China}}

\begin{abstract}
Retrieval-augmented generation (RAG) grounds large language models with external evidence, but under a limited context budget, the key challenge is deciding \emph{which} retrieved passages should be injected.
We show that retrieval relevance metrics (e.g., NDCG) correlate weakly with end-to-end QA quality and can even become negatively correlated under multi-passage injection, where redundancy and mild conflicts destabilize generation.
We propose \textbf{Information Gain Pruning (IGP)}, a deployment-friendly reranking-and-pruning module that selects evidence using a generator-aligned utility signal and filters weak or harmful passages before truncation, without changing existing budget interfaces.
Across five open-domain QA benchmarks and multiple retrievers and generators, IGP consistently improves the quality--cost trade-off.
In a representative multi-evidence setting, IGP delivers about \textbf{+12--20\%} relative improvement in average F1 while reducing final-stage input tokens by roughly \textbf{76--79\%} compared to retriever-only baselines.
\end{abstract}


\maketitle


\section{Introduction}
\label{sec:intro}

\noindent\textbf{Background.}
Large language models (LLMs) have shown strong performance on open-domain question answering and complex reasoning tasks \citep{deepseek-r1}. 
However, when the input does not contain sufficient supporting evidence, LLMs can still generate factually incorrect or inconsistent content \citep{deng2025influence, self-contradic}. 
Retrieval-augmented generation (RAG) mitigates this issue by retrieving external documents and conditioning generation on the retrieved evidence, making answers more traceable and verifiable \citep{rag, rag-reduce-hallucination, survey-on-hallucination-2023, survey-on-hallucination-2025}. 
In practical systems, a common pipeline is \texttt{retrieve $\rightarrow$ rerank $\rightarrow$ truncate}: the system first retrieves a set of candidate passages, then reranks them (typically by relevance), and finally truncates them under a fixed budget (e.g., \TopM\footnote{%
\TopM\ denotes the passage-budget interface for evidence injection: at most $M$ evidence passages are retained;
\TopK\ denotes that the generator returns only the top-$K$ token log-probabilities at each decoding step (used to compute the uncertainty proxy in this paper), to avoid confusion with the “top-$k$ retrieved documents” setting.} or a token budget) before injection \citep{nips-2024-rankrag, ke-etal-2024-bridging}.

\noindent\textbf{Motivation.}
In deployed RAG systems, the evidence-injection budget is often tight, making evidence selection under the budget a primary practical concern \citep{rag-vs-long-context}.
Injecting multiple pieces of evidence may introduce redundancy, ambiguity, and conflicting statements, which consume the context budget and disturb generation, causing unstable conclusions or unnecessary over-explanations \citep{lost-in-the-middle, jin-etal-2025-hierarchical}. More importantly, \textbf{relevance is not equivalent to the marginal utility of evidence for generation}. Even highly relevant passages may contain parallel claims, conditional constraints, or contradictory details, which can spread the generation distribution, making answers less stable and sometimes less accurate \citep{wang2025retrievalaugmented, jiang-etal-2025-gainrag}.

\begin{figure}[pos=htbp]
\centering
\includegraphics[width=0.9\linewidth]{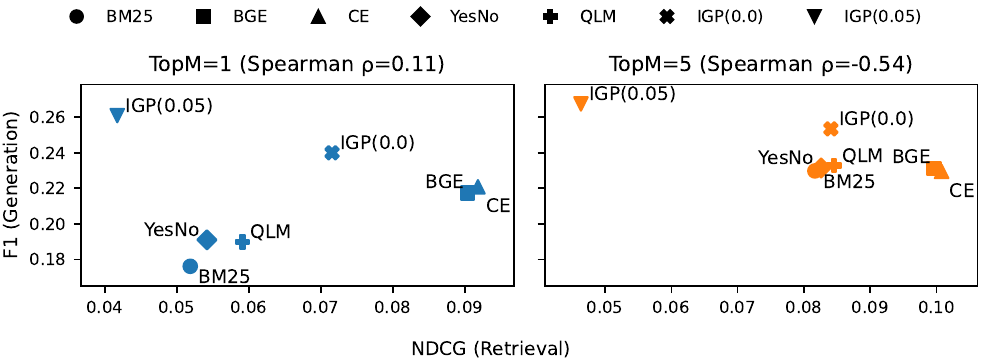}
\caption{\textbf{In RAG, better retrieval ranking does not necessarily yield better end-to-end generation.}
Under two evidence budgets (\TopM=1 and \TopM=5), the Spearman rank correlation between NDCG and end-to-end F1 is generally weak, and can be negative when multiple passages are injected (see \autoref{sec:relevance-vs-helpfulness}).}
\label{fig:retrieval-generation-corr}
\end{figure}

\noindent\textbf{First Contribution: Relevance–Utility Mismatch.}
We identify a persistent mismatch in practical budgeted RAG: \textbf{improving retrieval relevance does not reliably improve end-to-end answer quality}. Across evidence budgets, offline relevance metrics (e.g., NDCG) show weak and non-robust correlation with generation quality (F1), and the correlation can even turn negative when multiple passages are injected (e.g., \TopM$=5$) (\autoref{fig:retrieval-generation-corr}; see \autoref{sec:relevance-vs-helpfulness}). We attribute this to a generator-side mechanism: evidence injection reshapes token-level output distributions, and redundant, non-conclusive, or inconsistent passages can spread probability mass, increase uncertainty on key tokens, and destabilize generation under a fixed context budget. This motivates shifting the evidence-selection objective from “optimizing relevance” to “optimizing evidence utility for generation,” where good evidence should make the generator more concentrated and stable on key output tokens (i.e., reduce uncertainty).

\begin{figure}[pos=htbp]
\centering
\includegraphics[width=0.9\linewidth]{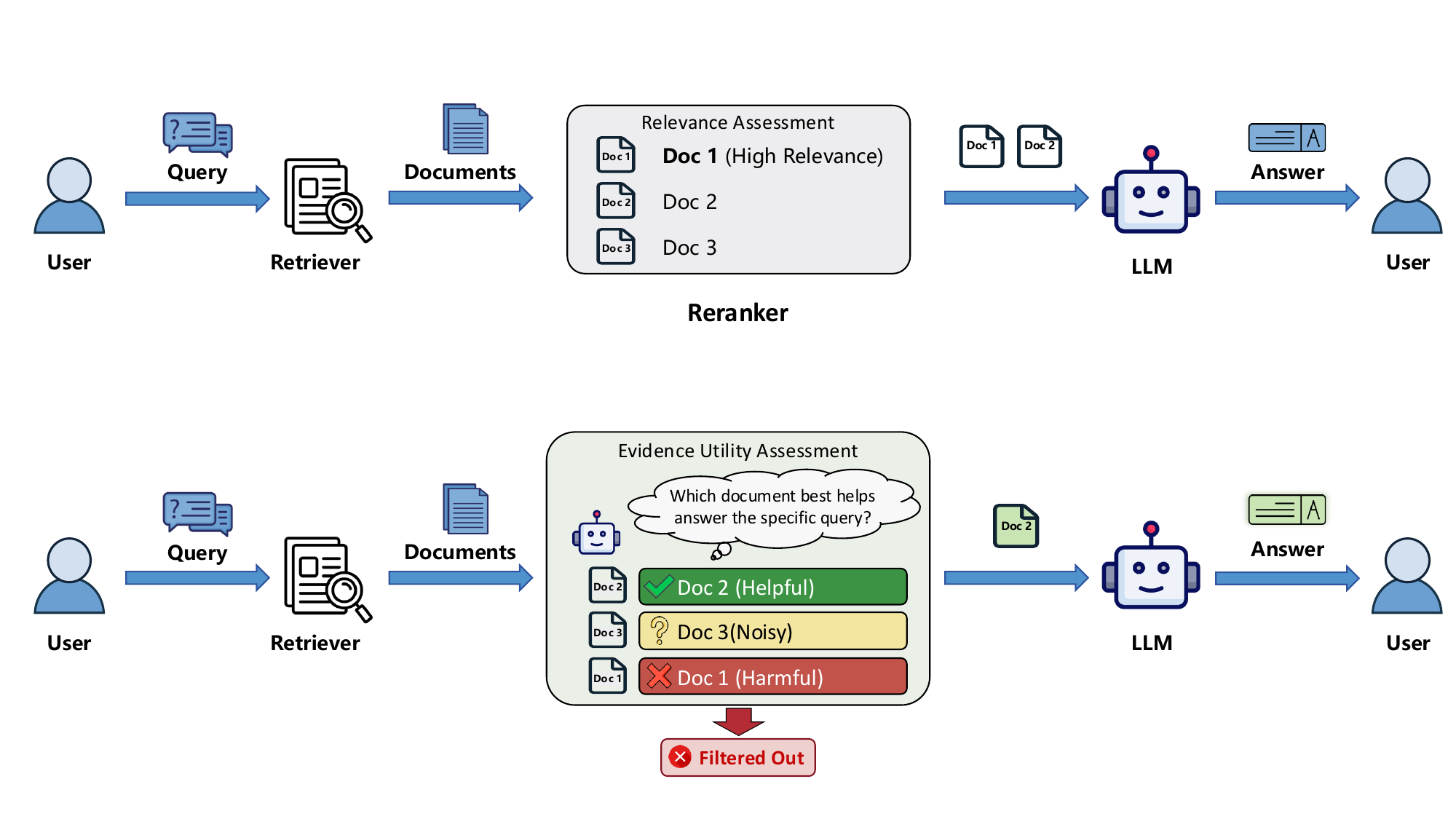}
\caption{\textbf{RAG pipeline comparison: relevance reranking vs.\ IGP.}
IGP replaces relevance scoring with an information-gain signal aligned with the generator, and can filter negative/weak-gain evidence; the budget-based truncation mechanism remains unchanged.}
\label{fig:rag-pipeline}
\end{figure}

\noindent\textbf{Second Contribution: Our Approach.}
We propose \textbf{Information Gain Pruning (IGP)} as a plug-and-play replacement for the \texttt{rerank} stage in \texttt{retrieve $\rightarrow$ rerank $\rightarrow$ truncate} (\autoref{fig:rag-pipeline}). IGP measures evidence utility from the generator’s perspective by constructing a normalized uncertainty (NU) from step-wise output distributions and defining information gain (IG) as the reduction in uncertainty after injecting a candidate passage. It then reranks candidates by IG and applies a threshold to prune negative-utility or weak-utility passages before truncation, preventing redundant or conflicting evidence from occupying the limited context. Importantly, IGP only replaces \texttt{rerank}; \texttt{truncate} and the original budget interface (\TopM or token budget) remain unchanged, and IGP requires \textbf{no labels, no training, and no parameter access}—only step-wise logits (or \TopK log-probabilities), making it compatible with black-box LLM services.

\noindent\textbf{Third Contribution: Main Findings.}
Across multiple open-domain QA benchmarks, different first-stage retrievers, and different generator families/scales, IGP consistently improves or maintains end-to-end answer quality \emph{while reducing the final-stage context cost}, thereby strengthening the Pareto frontier of quality versus cost.
Notably, in a representative \textbf{multi-evidence} setting, IGP delivers a clear \emph{win--win}: it improves the average token-level F1 by about \textbf{+12--20\%} while reducing the average final-stage input tokens by roughly \textbf{76--79\%} (relative to the retriever-only baseline).
Our analyses further show that these gains are driven primarily by \textbf{utility-aware admission control (threshold-based pruning)} rather than reordering alone, and are most pronounced when multiple passages are injected—where relevance-based reranking is more likely to admit redundant or mildly conflicting evidence under a fixed budget, amplifying generator uncertainty and hurting answer stability.
Together, these results suggest that \textbf{admission control guided by generator-aligned utility} is a key lever for more reliable and cost-effective budgeted RAG.

\noindent\textbf{Summary of Contributions.}
\begin{itemize}
    \item \textbf{Relevance--utility mismatch:}
    We show that offline relevance (e.g., NDCG) correlates weakly—and can be negative—with end-to-end QA quality (F1), especially under multi-evidence budgets.

    \item \textbf{IGP (generator-aligned rerank + pruning):}
    We propose Information Gain Pruning, which scores passages by uncertainty reduction (information gain) and prunes weak/negative-utility evidence before truncation. IGP is label-free, training-free, parameter-free, and requires only step-wise logits/\TopK outputs.

    \item \textbf{Better quality--cost frontier:}
    Across QA benchmarks, retrievers, and generator families/scales, IGP improves or maintains F1 while reducing final-stage input tokens, consistently strengthening the Pareto frontier—most notably in multi-evidence settings.
\end{itemize}

\noindent\textbf{Paper Organization.}
The remainder of this paper is organized as follows.
\autoref{sec:related-work} reviews related work on RAG, reranking/evidence selection, and uncertainty estimation for LLMs.
\autoref{sec:method} introduces Information Gain Pruning (IGP), including the \TopK normalized uncertainty proxy and the IG-based reranking-and-pruning algorithm.
\autoref{sec:experiments} presents the experimental setup and main results, followed by ablations and sensitivity analyses, the relevance--utility mismatch study, and robustness evaluations across retrievers and generator families/scales.
Finally, \autoref{sec:conclusion} concludes with practical implications, limitations, and future directions.

\section{Related Work}
\label{sec:related-work}

This paper targets a practical bottleneck in deployed RAG: under a limited context budget, the key question is \emph{which evidence should enter the generator context}, i.e., evidence that is \emph{most helpful for generation} rather than merely \emph{most relevant}. Related studies broadly fall into three directions: (1) RAG frameworks and retriever/reader design, (2) reranking and evidence selection in the \texttt{retrieve $\rightarrow$ rerank $\rightarrow$ truncate} pipeline, and (3) uncertainty quantification and hallucination detection for LLMs. Our Information Gain Pruning (IGP) sits at their intersection: it uses a generator-side uncertainty reduction signal to rerank candidates and filter out negative/weak-utility evidence \emph{before} truncation.

\subsection{Retrieval-Augmented Generation (RAG)}
Retrieval-Augmented Generation (RAG) augments LLMs with an external, updatable non-parametric corpus, so generation can be grounded in retrievable evidence and become more traceable. \citet{rag} propose the RAG framework that retrieves from large corpora (e.g., Wikipedia) and conditions generation on retrieved documents, improving factuality and interpretability on knowledge-intensive tasks.

On the retriever side, dense retrieval is a common front-end for open-domain QA and RAG. \citet{dpr} introduce DPR, a dual-encoder method that learns representations for queries and passages and significantly improves recall, making it a popular first-stage retriever. \citet{realm} further integrate retrieval into pre-training, enabling retrieval during pre-training, fine-tuning, and inference, and encouraging the model to learn how to use retrievable knowledge more systematically.

On the reader/generator side, \citet{fid} propose FiD, which separately encodes multiple evidence passages and fuses them in the decoder, strengthening multi-evidence aggregation and serving as a strong baseline in RAG-style open-domain QA. Overall, many systems follow the ``high-recall retrieval + strong reader/fusion'' paradigm. However, as emphasized in our introduction, under a fixed context budget (e.g., \TopM or token budget), multi-evidence injection often brings redundancy and conflicts, so retrieving ``relevant'' content alone does not guarantee better end-to-end generation.

\subsection{Reranking and Evidence Selection in RAG}
To decide what evidence enters the context, practical RAG systems commonly use \texttt{retrieve $\rightarrow$ rerank $\rightarrow$ truncate}: retrieve a candidate set, rerank by a scoring function (often relevance), and then truncate under a fixed budget (\TopM or token budget). A classic reranking approach is the discriminative cross-encoder. \citet{nogueira2019bert} jointly encode $(q,d)$ with BERT and output a relevance score, showing strong passage reranking performance, but with high inference cost, it is typically used as a second-stage reranker.

Generative reranking is another line. \citet{monot5} introduce MonoT5, which reformulates ranking as a sequence-to-sequence task and derives relevance scores from generated labels or target-token logits, achieving competitive ranking quality. To balance effectiveness and efficiency, \citet{colbert} propose ColBERT with late interaction, enabling scalable token-level matching while retaining strong performance.

More recently, LLMs have been explored as rerankers. \citet{rankgpt} study instruction-based LLM reranking (e.g., ChatGPT/GPT-4) and discuss prompting and distillation for efficiency and generalization. Despite these advances, most reranking methods still optimize \emph{relevance}. In contrast, our introduction highlights a relevance--utility mismatch: under limited budgets, especially when multiple passages are injected, ``more relevant'' evidence can still be redundant, non-conclusive, or conflicting, and thus may reduce the \emph{marginal utility} of the context for generation. This motivates evidence selection objectives that are aligned with the generator, not only retrieval relevance.

\subsection{Uncertainty Quantification and Hallucination Detection for LLMs}
LLMs can be fluent but unreliable when evidence is insufficient or inconsistent, making uncertainty quantification (UQ) important for trustworthy systems. \citet{shorinwa2025survey} provide a comprehensive survey of LLM uncertainty methods, covering white-box signals (e.g., token probabilities, entropy) and black-box signals (e.g., multi-sample consistency and semantic grouping), and summarize their use in hallucination detection and risk-aware decision making.

Calibration studies the match between confidence and true correctness. \citet{guo2017calibration} show that modern neural models are often overconfident and propose post-hoc calibration methods such as temperature scaling. For epistemic uncertainty, common approximations include MC dropout and deep ensembles: \citet{gal2016dropout} interpret dropout as approximate Bayesian inference, while \citet{lakshminarayanan2017ensembles} show deep ensembles can provide strong and robust uncertainty estimates.

In closed-source or black-box LLM settings, full distributions and internal states are usually unavailable. \citet{xiong2024can} propose a black-box framework combining verbalized confidence prompts with multi-sample consistency and aggregation, showing that self-reported confidence can be miscalibrated but sampling-based signals can help predict failures. Uncertainty is also used to detect hallucinations: \citet{semanticentropy} propose semantic entropy by sampling multiple answers and clustering them by meaning; larger semantic disagreement indicates higher hallucination risk and can support refusal, triggering retrieval, or human review.

\subsection{Positioning of This Paper}
Prior work makes progress in RAG, reranking, and UQ, but they are often used in a loosely connected way: RAG focuses on retrieving evidence, reranking focuses on relevance ordering, and UQ is usually applied \emph{after} generation for confidence reporting or hallucination detection. Our \textbf{IGP} differs in three aligned aspects:

\begin{itemize}
  \item \textbf{From relevance to evidence utility for generation:} Instead of ranking evidence by relevance, IGP uses a generator-aligned signal---whether evidence injection reduces generation uncertainty---to estimate the \emph{marginal utility} of a passage under a fixed context budget.
  \item \textbf{Earlier control at the \texttt{rerank} stage:} Rather than detecting issues after an answer is produced, IGP performs \emph{utility-aware reranking and threshold-based pruning} before evidence enters the context, filtering redundant, noisy, or conflicting passages that provide negative/weak gain.
  \item \textbf{Deployment-friendly and black-box compatible:} Consistent with our introduction, IGP is label-free, training-free, and parameter-free. It only needs step-wise logits or \TopK log-probabilities, and it \emph{replaces only \texttt{rerank}} while keeping \texttt{truncate} and the original budget interface (e.g., \TopM or token budget) unchanged.
\end{itemize}

In short, this paper moves uncertainty from an \emph{answer-level} diagnostic to an \emph{evidence-level} decision signal, offering a modular, generator-aligned method for evidence admission under budget-limited RAG deployment.

\section{Methodology}
\label{sec:method}

This section introduces Information Gain Pruning (IGP) and its uncertainty-reduction signal used for evidence selection in budgeted RAG.
We first give an overview of the pipeline replacement and key quantities, then define normalized uncertainty (NU), information gain (IG), and the final reranking-and-pruning algorithm, followed by practical implementation notes.

\subsection{Overview}
\label{sec:method_overview}

\paragraph{Problem setting.}
We consider a standard budgeted RAG pipeline,
\texttt{retrieve $\rightarrow$ rerank $\rightarrow$ truncate},
where a first-stage retriever returns candidates $\mathcal{D}$ and the system must select evidence for the generator under a fixed budget (e.g., Top-$M$, optionally with a token guard $B$).
In this regime, relevance-based reranking can be suboptimal: passages may be relevant yet redundant, non-decisive, or mutually inconsistent, consuming the limited context and destabilizing generation.

\paragraph{Key idea.}
We align evidence selection with the generator rather than retrieval relevance.
Our intuition is that \emph{useful} evidence should make the model more decisive under a fixed decoding protocol, i.e., it should \emph{reduce} token-level uncertainty over the next-token distribution.
This yields an evidence-level criterion: a passage is helpful if injecting it decreases the generator's uncertainty compared to using no evidence.

\paragraph{Information Gain Pruning (IGP).}
We propose \textbf{Information Gain Pruning (IGP)} as a drop-in replacement for \texttt{rerank}:
\[
\texttt{retrieve} \;\rightarrow\; \texttt{IGP} \;\rightarrow\; \texttt{truncate}.
\]
IGP computes a black-box uncertainty proxy from step-wise outputs using only Top-$K$ log-probabilities, forming \textbf{normalized uncertainty (NU)}.
It then defines a passage utility score, \textbf{information gain (IG)}, as the reduction in NU after injecting the passage.
Candidates are ranked by IG and filtered by a threshold $T_p$ to prune weak or negative-utility evidence \emph{before} truncation, preventing low-gain passages from occupying the limited context.

\paragraph{Interface and requirements.}
IGP is \textbf{label-free, training-free, and parameter-free}.
It requires only \textbf{black-box} access to step-wise logits (or Top-$K$ log-probabilities) under a fixed prompt and a deterministic probing protocol.
Importantly, IGP changes only the \texttt{rerank} stage; the downstream \texttt{truncate} module and the original budget interface (Top-$M$ and optional $B$) remain unchanged, ensuring plug-and-play compatibility with existing systems.

\paragraph{Notation.}
Frequently used symbols are summarized in \autoref{tab:notation_method}, with key quantities defined in \autoref{sec:method_nu} and \autoref{sec:method_ig}.

\begin{table}[pos=htbp]
\centering
\small
\setlength{\tabcolsep}{6pt}
\begin{tabular}{ll}
\toprule
\textbf{Symbol} & \textbf{Meaning} \\
\midrule
$q$ & Query/question. \\
$\mathcal{D}=\{d_i\}_{i=1}^{N}$ & First-stage retrieved candidate evidence passages. \\
$\mathcal{L}$ & Candidate list output by IGP after reranking and pruning. \\
$\mathcal{S}$ & Final retained evidence set after \texttt{truncate}. \\
$\phi$ & Generator (black-box; returns logits or Top-$K$ log-probabilities). \\
$\mathcal{V}$, $V$ & Vocabulary and its size. \\
$\mathbf{z}_t \in \mathbb{R}^{V}$ & Logits at decoding step $t$. \\
$\hat{y}_t$ & Greedy token at step $t$. \\
$K$ & Top-$K$ truncation hyperparameter for NU/IG estimation. \\
$\mathcal{V}^{(K)}_t$ & Top-$K$ token set at step $t$. \\
$MT$ & Maximum rollout length for uncertainty estimation. \\
$T$ & Effective rollout length (EOS or $MT$ truncation). \\
$M$ & Evidence budget (Top-$M$): number of retained passages (enforced by \texttt{truncate}). \\
$B$ & (Optional) token-budget guard to satisfy hard input-length constraints. \\
$\widehat{NU}(q;\phi,K)$ & Unconditional normalized uncertainty. \\
$\widehat{NU}(q\mid d;\phi,K)$ & Conditional normalized uncertainty given a passage $d$. \\
$IG(d,q;\phi,K)$ & Information gain / utility proxy: uncertainty reduction. \\
$T_p$ & Pruning threshold (admission barrier). \\
\bottomrule
\end{tabular}
\caption{Main notation used in \autoref{sec:method}.}
\label{tab:notation_method}
\end{table}

\subsection{Uncertainty Proxy: \TopK Normalized Uncertainty (NU)}
\label{sec:method_nu}

\paragraph{Deterministic probing protocol and effective length.}
To obtain a reproducible and low-variance signal, we use greedy decoding (temperature $0$) when estimating NU/IG.
At step $t$, the greedy token is $\hat{y}_t=\arg\max_{v\in\mathcal{V}} z_{t,v}$.
Generation stops when EOS is produced or when reaching the maximum rollout length $MT$.
We define the effective rollout length as:
\begin{equation}
T=\min\{MT,\ T_{\mathrm{EOS}}\},\qquad
T_{\mathrm{EOS}}=\min\{t:\hat{y}_t=\mathrm{EOS}\}.
\label{eq:mt_trunc_rule_method}
\end{equation}

\paragraph{\TopK truncation and renormalization.}
Let $\mathcal{V}^{(K)}_t$ be the \TopK token set at step $t$.
We renormalize the probability mass only over this set:
\begin{equation}
\tilde{p}_\phi(y_t=v\mid \hat{y}_{<t},q;K)
=
\frac{\exp(z_{t,v})}{\sum_{v'\in \mathcal{V}^{(K)}_t}\exp(z_{t,v'})},
\quad
v\in\mathcal{V}^{(K)}_t.
\label{eq:topk_softmax_method}
\end{equation}

\paragraph{Token-level entropy and normalized uncertainty.}
We compute the \TopK entropy and its normalized form:
\begin{equation}
\tilde{H}_t(q;\phi,K)
=
-\sum_{v\in \mathcal{V}^{(K)}_t}
\tilde{p}_\phi(v\mid \hat{y}_{<t},q;K)\,
\log \tilde{p}_\phi(v\mid \hat{y}_{<t},q;K),
\label{eq:topk_token_entropy_method}
\end{equation}
\begin{equation}
\tilde{u}_t(q;\phi,K)=\frac{\tilde{H}_t(q;\phi,K)}{\log K}\in[0,1].
\label{eq:topk_normalized_uncertainty_method}
\end{equation}

\paragraph{Sequence-level uncertainty.}
We define a sequence-level normalized uncertainty by averaging across steps:
\begin{equation}
\widehat{NU}(q;\phi,K)
=
\frac{1}{T}\sum_{t=1}^{T}\tilde{u}_t(q;\phi,K)
=
\frac{1}{T\log K}\sum_{t=1}^{T}\tilde{H}_t(q;\phi,K).
\label{eq:nu_hat_uncond_method}
\end{equation}

\paragraph{Single-evidence conditional uncertainty.}
After injecting a single candidate passage $d$ (through a fixed prompt interface), we obtain a rollout $\hat{y}^{(d)}$ and length $T^{(d)}$.
We define the conditional normalized uncertainty as:
\begin{equation}
\widehat{NU}(q\mid d;\phi,K)
=
\frac{1}{T^{(d)}\log K}\sum_{t=1}^{T^{(d)}}
\Bigg(
-\sum_{v\in \mathcal{V}^{(K,d)}_t}
\tilde{p}_\phi(v\mid \hat{y}^{(d)}_{<t},q,d;K)\,
\log \tilde{p}_\phi(v\mid \hat{y}^{(d)}_{<t},q,d;K)
\Bigg).
\label{eq:nu_hat_cond_method}
\end{equation}

\subsection{Information Gain (IG): A Generator-Aligned Evidence-Utility Signal}
\label{sec:method_ig}

\paragraph{Definition.}
We define \emph{evidence utility for generation} (information gain) as the reduction in uncertainty:
\begin{equation}
IG(d,q;\phi,K)
=
\widehat{NU}(q;\phi,K)-\widehat{NU}(q\mid d;\phi,K).
\label{eq:ig_def_method}
\end{equation}
$IG>0$ means the passage makes generation more certain (uncertainty decreases), while $IG<0$ suggests the passage may introduce noise or conflicts that spread the output distribution and reduce utility under a limited context budget.

\paragraph{We target ranking consistency, not an unbiased information-theoretic estimate.}
We do not treat \eqref{eq:ig_def_method} as an unbiased estimator of true conditional mutual information.
Our goal is a practical proxy that is \emph{reproducible} and \emph{low-variance} under a fixed prompt interface and a fixed decoding protocol, so that the \emph{relative ordering} of candidates matches their end-to-end utility for generation.

\paragraph{Why Top-$K$ entropy under greedy decoding helps utility ranking.}
Under a fixed prompt and greedy decoding, a high-utility passage often increases the logit margin of the intended next tokens at multiple key steps.
This makes the Top-$K$ probability mass more concentrated and reduces Top-$K$ entropy.
Because $\widehat{NU}$ averages entropy across steps (Eq.~\eqref{eq:nu_hat_uncond_method}--\eqref{eq:nu_hat_cond_method}), consistent entropy reductions over several key steps accumulate into a stable passage-level difference.
In contrast, passages that introduce competing hypotheses tend to flatten probabilities among alternative tokens, increasing entropy and producing smaller (or even negative) $IG$.
This is an intuition for rank preservation, rather than a formal guarantee.

\subsection{IGP: Utility-Aware Reranking and Pruning, While Keeping the \TopM Budget Interface}
\label{sec:method_igp}

IGP computes $IG$ for each candidate passage and produces an output list $\mathcal{L}$.
It first sorts candidates by $IG$ in descending order and then applies a pruning threshold $T_p$ to filter weak or negative-utility evidence (keeping only those with $IG(d,q;\phi,K)\ge T_p$).
This provides explicit evidence admission control and improves the quality of the prefix under a limited context budget.
After that, the original \texttt{truncate} module applies the \TopM rule (and optionally the token guard $B$) on $\mathcal{L}$ to obtain the final injected set $\mathcal{S}$.

\begin{algorithm}[htbp]
\caption{IGP as a plug-and-play replacement for rerank: retrieve-IGP-truncate}
\label{alg:igp_method}
\begin{algorithmic}[1]
\Require Query $q$; candidates $\mathcal{D}=\{d_i\}_{i=1}^{N}$; generator $\phi$;
\TopK parameter $K$; maximum rollout length $MT$;
pruning threshold $T_p$;
budget interface: \TopM (optional token guard $B$).
\Ensure Threshold-filtered and reranked candidate list $\mathcal{L}$; final evidence set $\mathcal{S}$ (produced by \texttt{truncate}).

\State $\nu_0 \gets \widehat{NU}(q;\phi,K)$ \Comment{computed once}
\ForAll{$i=1$ to $N$ \textbf{in parallel}} \Comment{batch/parallel rollouts over passages}
  \State $\nu_i \gets \widehat{NU}(q\mid d_i;\phi,K)$
  \State $s_i \gets \nu_0 - \nu_i$
\EndFor
\State $\pi \gets \mathrm{argsort}_i(s_i;\downarrow)$
\State $\mathcal{L} \gets$ the list obtained by ordering $\mathcal{D}$ with $\pi$
\State $\mathcal{L} \gets [\,d \in \mathcal{L} \;:\; IG(d,q;\phi,K)\ge T_p\,]$
\State $\mathcal{S} \gets \mathrm{Truncate}(\mathcal{L};\,M,\ B)$ \Comment{\TopM (optional token guard)}
\State \Return $\mathcal{L}, \mathcal{S}$
\end{algorithmic}
\end{algorithm}

\paragraph{Scope: IGP vs.\ \texttt{truncate}.}
IGP decides \emph{which passages are admitted and how they are ordered} (reranking + pruning), but it does not decide \emph{how many} passages are ultimately injected.
The final evidence size is still enforced by \texttt{truncate} through the same \TopM interface (and optionally the token guard $B$).
In many RAG corpora, passages are chunked into roughly similar lengths, so Top-$M$ is a practical approximation of a fixed token budget.
When passage lengths vary substantially, enabling the token-budget guard $B$ inside \texttt{truncate} satisfies hard length constraints without changing the definition or scope of IGP.

\subsection{Implementation Notes}
\label{sec:method_practical}

\paragraph{Black-box interface assumption.}
IGP does not require gradients or model parameters, and it uses no labels or training.
Its only requirement is that the generator $\phi$, under a fixed prompt interface and decoding protocol, can return step-wise logits (or equivalently \TopK log-probabilities), which are used to compute \TopK entropy and $\widehat{NU}$.

\paragraph{Deterministic probing vs.\ answer generation.}
We use greedy decoding (temperature $0$) only when estimating NU/IG, because it makes the signal reproducible and reduces variance.
This probing protocol is used to compare candidate passages and does not change the system's budget interface.
In practice, the final answer generation stage can follow the original decoding setting of the deployed system.

\paragraph{Meaning of the threshold $T_p$ (admission barrier).}
The threshold $T_p$ controls evidence admission: only passages with $IG(d,q;\phi,K)\ge T_p$ enter $\mathcal{L}$ and participate in later \TopM truncation.
A larger $T_p$ is more conservative and filters more weak/negative-utility passages, while a smaller $T_p$ keeps more candidates to preserve coverage.
To approximate ``no pruning,'' one can set $T_p$ to a sufficiently small value (e.g., $T_p=-\infty$ or a small negative number), in which case the interface and behavior reduce to pure IG-based reranking followed by the original \texttt{truncate} module (i.e., the system retains the same \TopM/\,$B$ budget contract, only changing the ordering signal).

\paragraph{Choice of $K$ and $MT$.}
$K$ controls the granularity of the uncertainty proxy.
A very small $K$ can make $\tilde{H}_t$ overly sensitive to local logit ties, while an excessively large $K$ may include many low-probability tail tokens that are noisy under black-box \TopK access.
In practice, we treat $K$ as an estimation knob: moderate values typically yield stable rankings because the entropy mass is dominated by the head of the distribution under greedy decoding.
The maximum rollout length $MT$ bounds computation and standardizes the probing horizon.
We use the same $MT$ for the unconditional rollout and all conditional rollouts to ensure comparability (Eq.~\eqref{eq:mt_trunc_rule_method}).

\paragraph{Length effects and fair comparison across passages.}
Because injecting evidence can change the effective length ($T^{(d)}$) by triggering earlier/later EOS, we define $\widehat{NU}$ as a per-step average (Eq.~\eqref{eq:nu_hat_uncond_method}--\eqref{eq:nu_hat_cond_method}), which normalizes by the realized rollout length.
This makes $\widehat{NU}$ comparable even when $T^{(d)}$ varies.
If desired, one can further enforce a common horizon by truncating all rollouts to $\min(T, T^{(d)})$ when computing the sums, but we find the length-normalized definition already provides a low-variance ranking signal under deterministic probing.

\paragraph{Prompting convention and evidence injection.}
IGP treats the generator $\phi$ as a black box; the only requirement is a fixed prompt interface that deterministically injects a single passage $d$ (e.g., via a dedicated \texttt{Evidence:} field) while keeping the query $q$ and all other formatting identical.
To avoid confounding factors, the system prompt, decoding protocol (temperature $0$ for probing), and any tool or formatting instructions are held constant across all rollouts.
Only the presence/absence of the candidate passage differs between $\widehat{NU}(q;\phi,K)$ and $\widehat{NU}(q\mid d;\phi,K)$.

\paragraph{Computation, batching, and caching.}
IGP requires one unconditional rollout and $N$ conditional rollouts per query, i.e., $N{+}1$ probing calls to $\phi$.
This cost is embarrassingly parallel across passages (Alg.~\ref{alg:igp_method}, line 2), and can be amortized in practice via batching over candidates and caching the unconditional baseline $\nu_0$ (computed once per query).
When the generator API returns only \TopK log-probabilities, NU/IG computation remains feasible and efficient, since Eq.~\eqref{eq:topk_softmax_method}--\eqref{eq:topk_token_entropy_method} use only the \TopK set and renormalization.

\paragraph{Multi-evidence interactions and scope.}
IGP evaluates each passage independently via single-evidence uncertainty reduction (Eq.~\eqref{eq:ig_def_method}).
This design keeps the method label-free and plug-and-play, but it does not explicitly model higher-order interactions (e.g., two passages that are individually helpful but redundant together, or jointly needed for a key inference).
In our setting, we rely on the downstream \texttt{truncate} budget interface (\TopM and optional $B$) plus the admission barrier $T_p$ to mitigate obvious low-utility evidence before truncation, while preserving the simplicity and reproducibility of the generator-aligned scoring signal.

\section{Experiments}
\label{sec:experiments}

We evaluate \textbf{Information Gain Pruning (IGP)} as a drop-in replacement of the \texttt{rerank} stage in a budgeted RAG pipeline (\texttt{retrieve $\rightarrow$ rerank $\rightarrow$ truncate}). We keep the corpus, candidate size, prompts/decoding, and the evidence-budget executor fixed across methods, and vary only the reranking/admission strategy to isolate the end-to-end effects.

\paragraph{Research Questions (RQs).}
\begin{itemize}
    \item \textbf{RQ1 (Quality--cost trade-off):}
    Does IGP improve the end-to-end quality--cost frontier under different evidence budgets?
    \textbf{Answered in \autoref{sec:main_results_bm25}.}

    \item \textbf{RQ2 (Mechanism and sensitivity):}
    Where do the gains come from (reordering vs.\ admission control), and how sensitive is IGP to the pruning threshold and uncertainty-estimation hyperparameters?
    \textbf{Answered in \autoref{sec:ablation_results}.}

    \item \textbf{RQ3 (Relevance $\neq$ utility):}
    How well do offline relevance metrics (NDCG) predict end-to-end QA quality (F1) under different budgets?
    \textbf{Answered in \autoref{sec:relevance-vs-helpfulness}.}

    \item \textbf{RQ4 (Retriever robustness):}
    Do the gains of IGP persist when changing the first-stage retriever?
    \textbf{Answered in \autoref{sec:retriever_effect}.}

    \item \textbf{RQ5 (Generator robustness):}
    Does IGP generalize across generator families and parameter scales?
    \textbf{Answered in \autoref{sec:model_scale}.}
\end{itemize}

\subsection{Experimental Setup}
\label{sec:exp_setup}

This subsection details the corpus, datasets, generator backbones, prompting/decoding protocols, baselines, and metrics. For fair comparison, we hold the pipeline fixed—same corpus, candidate size N, prompts, decoding settings, and budget executor (\texttt{truncate}: Top-M, optionally with token guard B)—and vary only the \texttt{rerank}/admission strategy (and M in budget sweeps). Unless specified, IGP is a plug-and-play replacement for \texttt{rerank}; RQ4 additionally varies the first-stage retriever, and RQ5 varies the generator family/scale.

\subsubsection{Corpus}
\label{sec:corpus}

We use the Wikipedia retrieval corpus released by FlashRAG \citep{flashrag} as the external knowledge base\footnote{\url{https://huggingface.co/datasets/RUC-NLPIR/FlashRAG_datasets/blob/main/retrieval-corpus/wiki18_100w.zip}}.
This corpus is built from the English Wikipedia snapshot dated 2018-12-20.
After cleaning, it is segmented into non-overlapping passages of \textbf{100 words}, resulting in about \textbf{21M} passages.
Unless otherwise stated, all retrieval, reranking, and truncation are conducted on this corpus.

\subsubsection{Datasets}
\label{sec:datasets}

\begin{table}[pos=htbp]
\centering
\small
\begin{tabular}{lrrrl}
\hline
\textbf{Dataset} & \textbf{\#Train} & \textbf{\#Dev} & \textbf{\#Test} & \textbf{Category} \\
\hline
NQ        & 79,168 & 8,757  & \textbf{3,610}  & Real queries \\
TriviaQA  & 78,785 & 8,837  & \textbf{11,313} & Trivia facts \\
PopQA     & /      & /      & \textbf{14,267} & Long-tail knowledge \\
SQuAD     & 87,599 & \textbf{10,570} & /      & Reading comprehension \\
AmbigQA   & 10,036 & \textbf{2,002}  & /      & Disambiguation \\
\hline
\end{tabular}
\caption{Dataset statistics and category labels (\textbf{bold} indicates the split used in our evaluation).}
\label{tab:datasets}
\end{table}

We evaluate on five classic QA benchmarks with complementary properties:
\textbf{NQ} \citep{nq} (real-world search queries),
\textbf{TriviaQA} \citep{triviaqa} (open-domain factual QA),
\textbf{PopQA} \citep{popqa} (popular culture and long-tail entities),
\textbf{SQuAD} \citep{squad} (reading comprehension / extractive-style),
and \textbf{AmbigQA} \citep{ambigqa} (open-domain QA with \textbf{question ambiguity}, requiring multiple valid interpretations/answers).
As shown in \autoref{tab:datasets}, we report results on the \textbf{test} split by default; when the test split is not publicly available, we use the \textbf{dev} split (thus NQ/TriviaQA/PopQA use test, and SQuAD/AmbigQA use dev).

\subsubsection{Language Models (Generator Backbones)}
\label{sec:llm_backbones}

To validate the \textbf{black-box usability} and \textbf{model-agnostic} nature of IGP, and to systematically study the effect of \emph{model family} and \emph{parameter scale} on end-to-end RAG (see \autoref{sec:model_scale}),
we choose two mainstream open-source instruction-tuned model families as generators:
\textbf{Qwen2.5-Instruct} and \textbf{Llama-3.x-Instruct}.
All models are from the Hugging Face (HF) hub.
Inference is implemented with \textbf{vLLM}, and we export step-wise logits (or equivalently \TopK log-probabilities) without gradients or fine-tuning.
These outputs are used to compute \TopK normalized uncertainty and information gain (IG), which drives IGP reranking and pruning.

\paragraph{Qwen2.5 family (main experiments and scale ablation)}
Qwen2.5 provides instruction models across multiple scales\footnote{\url{https://huggingface.co/collections/Qwen/qwen25}}.
Unless otherwise specified, our main setting uses:
\begin{itemize}
    \item \textbf{Main:} \texttt{Qwen/Qwen2.5-7B-Instruct}.
\end{itemize}
For scale effects (\autoref{sec:model_scale}), we additionally evaluate:
\begin{itemize}
    \item \textbf{Scales:} \texttt{Qwen/Qwen2.5-0.5B-Instruct},
    \texttt{Qwen/Qwen2.5-1.5B-Instruct},
    \texttt{Qwen/Qwen2.5-3B-Instruct},
    \texttt{Qwen/Qwen2.5-7B-Instruct}.
\end{itemize}

\paragraph{Llama 3.x family (cross-family generalization and scale ablation)}
To test whether IGP generalizes across model families, we include Meta's Llama 3.x instruction models\footnote{\url{https://huggingface.co/meta-llama}}, covering small to medium sizes:
\begin{itemize}
    \item \textbf{Scales:} \texttt{meta-llama/Llama-3.2-1B-Instruct},
    \texttt{meta-llama/Llama-3.2-3B-Instruct},
    \texttt{meta-llama/Llama-3.1-8B-Instruct}.
\end{itemize}

\subsubsection{Prompts and Decoding Protocol}
\label{sec:prompts}

To ensure fair end-to-end comparisons across different retrieval/reranking/truncation strategies, we fix the prompt templates for all experiments.
We use two prompts: \textbf{(A) the final RAG answer-generation prompt} and \textbf{(B) the NU/IG probing prompt}.
Prompt (A) enforces evidence grounding for end-to-end QA, while prompt (B) is short and neutral to obtain a more reliable logit-change signal with low prompt-induced disturbance.
Except for document ordering/selection, all other input formats remain the same.

\paragraph{(A) Final RAG answer-generation prompt (grounded QA)}
This stage requires the model to answer using only the provided documents and \textbf{output only the answer}.
Let the concatenated documents be \texttt{\{reference\}} and the question be \texttt{\{question\}}:
\begin{quote}\small
\textbf{System:} You are given a question and a set of documents.\\
Answer the question using \textbf{only} the information in the documents.\\
Output \textbf{only the answer}.\\
\\
\textbf{User:} Documents:\\
\{reference\}\\
\\
Question: \{question\}\\
Answer:
\end{quote}

\noindent\textbf{Document concatenation format.}
\begin{quote}\small
\texttt{[DOC 1]} \{passage\_1\}\\
\texttt{[DOC 2]} \{passage\_2\}\\
$\cdots$\\
\texttt{[DOC M]} \{passage\_M\}
\end{quote}
Here $M$ is the number of retained passages, determined by the truncation depth or budget.
All methods share this format; they only differ in passage ordering and filtering.

\paragraph{(B) NU/IG probing prompt (short, low disturbance)}
NU/IG aims to estimate distribution changes before and after injecting a passage, so we use a more neutral and short template to reduce prompt effects on logits.
The unconditional template ($d=\varnothing$) and the single-passage conditional template ($d\neq\varnothing$) are:
\begin{quote}\small
\textbf{(1) Unconditional (}d=$\varnothing$\textbf{):}\\
\textbf{User:} \{question\}\\
\textbf{Assistant:}
\end{quote}

\begin{quote}\small
\textbf{(2) Conditional (}d$\neq\varnothing$\textbf{):}\\
\textbf{User:} \{question\}\\
\textbf{Context:}\\
\{context\}\\
\textbf{Assistant:}
\end{quote}
Template (1) is used to estimate $\widehat{NU}(q;\phi,K)$, and template (2) is used to estimate $\widehat{NU}(q\mid d;\phi,K)$ and compute $IG(d,q;\phi,K)$.

\paragraph{Unified implementation conventions.}
All methods use the same evidence injection format and the same \texttt{truncate} executor.
Except for hyperparameter sensitivity analyses in \autoref{sec:topk_mt_sensitivity}, we use \TopK=128 and $MT=32$ for NU/IG estimation.

\subsubsection{Baselines: Representative Retrievers, Rerankers, and Scoring Signals}
\label{sec:baselines}

To isolate where IGP gains come from, we compare against a set of \textbf{complementary and representative} baselines along common RAG evidence-selection axes: \emph{retrieval paradigm}, \emph{reranker architecture}, \emph{supervision source}, and \emph{scoring semantics}.
All methods are compared under the same end-to-end interface:
\[
\texttt{retrieve}\ \rightarrow\ \texttt{rerank}\ \rightarrow\ \texttt{truncate},
\]
and we \textbf{use \TopM as the evidence-injection budget interface}.
Since our passages are chunked with similar lengths, \TopM is a practical proxy for a token budget; thus differences mainly come from the \texttt{rerank} stage.

\begin{table}[pos=htbp]
\centering
\small
\setlength{\tabcolsep}{5pt}
\begin{tabular}{p{1.5cm}p{1.8cm}p{5.6cm}p{5.5cm}}
\toprule
\textbf{Method} & \textbf{Module} & \textbf{Architecture / score} & \textbf{Training / supervision} \\
\midrule
BM25 & Retrieve & Sparse term matching & Unsupervised (heuristic) \\
Contriever\footnote{\url{https://huggingface.co/facebook/contriever}} & Retrieve & Bi-encoder dense retrieval & Unsupervised / contrastive pretraining \\
BGE\footnote{\url{https://huggingface.co/BAAI/bge-reranker-base}} & Rerank & Bi-encoder similarity re-scoring & Supervised contrastive learning \\
CE\footnote{\url{https://huggingface.co/cross-encoder/ms-marco-MiniLM-L12-v2}} & Rerank & Cross-encoder relevance scoring & Fine-tuned with relevance labels \\
QLM\footnote{QLM uses the generator as a scorer; default generator is \texttt{Qwen/Qwen2.5-7B-Instruct}: \url{https://huggingface.co/Qwen/Qwen2.5-7B-Instruct}} & Rerank & Generative $p(q\mid d)$ (Query Likelihood) & Zero-shot (no task fine-tuning) \\
YesNo\footnote{YesNo uses the generator as a scorer; default generator is \texttt{Qwen/Qwen2.5-7B-Instruct}:
\url{https://huggingface.co/Qwen/Qwen2.5-7B-Instruct}} & Rerank & Discriminative $\Pr(\texttt{Yes}\mid q,d)$ & Zero-shot (no task fine-tuning) \\
IGP (ours) & Rerank & logits$\rightarrow$NU$\rightarrow$IG (uncertainty reduction) & Zero-shot (no task fine-tuning) \\
\bottomrule
\end{tabular}
\caption{Overview of retrievers, rerankers, and IGP used in this paper (HF links are provided in footnotes).}
\label{tab:retrievers_rerankers}
\end{table}

\paragraph{Baselines and references.}
We compare against sparse BM25~\citep{bm25} and unsupervised dense retrieval with Contriever~\citep{contriever}.
For \texttt{rerank}, we consider (i) supervised relevance rerankers: a lightweight bi-encoder reranker (BGE)~\citep{bge} and a cross-encoder reranker (CE; MiniLM fine-tuned on MS MARCO)~\citep{minilm,msmarco}; and (ii) generator-as-a-scorer baselines: query likelihood modeling (QLM)~\citep{qlm} and a binary Yes/No discriminative prompt score (YesNo)~\citep{yesno}.
Finally, IGP (ours) provides a black-box, generator-aligned \emph{utility} signal based on uncertainty reduction.

\subsubsection{Evaluation Metrics and Cost Accounting}
\label{sec:metrics}

We evaluate end-to-end RAG from three aspects: \textbf{(i) answer quality}, \textbf{(ii) inference cost}, and \textbf{(iii) cost-efficiency}.
Unless otherwise specified, all metrics are averaged over samples at the dataset level, where $N$ denotes the number of evaluation samples and $i\in\{1,\ldots,N\}$ indexes a sample.

\paragraph{Answer Quality: token-level $F_1$ (F1).}
We use token-level Precision/Recall/$F_1$, a standard metric in open-domain QA.
Let $\hat{y}$ be the predicted answer string, and let $\mathcal{Y}=\{y^{(1)},\dots,y^{(m)}\}$ be the set of $m$ reference answers for the same question.
We first apply a normalization function $\mathrm{norm}(\cdot)$ (e.g., lowercasing and removing punctuation/extra spaces), then split by whitespace into token \emph{multisets} via $\mathrm{split}(\cdot)$.
Formally, the token multiset of the prediction and the $j$-th reference are defined as:
\begin{equation}
\label{eq:tok_multisets}
T(\hat{y})=\mathrm{split}(\mathrm{norm}(\hat{y})),\qquad
T(y^{(j)})=\mathrm{split}(\mathrm{norm}(y^{(j)})).
\end{equation}
Here, $T(\cdot)$ returns a multiset of tokens, and $|T(\cdot)|$ denotes the multiset cardinality (i.e., total token count, including repetitions).

To compute multiset overlap, let $\mathcal{V}$ be the token vocabulary under $\mathrm{split}(\cdot)$, and let $c(t,T)$ be the multiplicity (count) of token $t\in\mathcal{V}$ in multiset $T$.
We define the number of overlapping tokens between prediction and reference as:
\begin{equation}
\label{eq:overlap}
\mathrm{overlap}\!\bigl(T(\hat{y}),T(y^{(j)})\bigr)
=\sum_{t\in\mathcal{V}} \min\!\Bigl(c\!\bigl(t,T(\hat{y})\bigr),\,c\!\bigl(t,T(y^{(j)})\bigr)\Bigr).
\end{equation}

Given $\mathrm{overlap}$, we define token-level precision $P^{(j)}$, recall $R^{(j)}$, and $F_1^{(j)}$ for the $j$-th reference. If there is no token overlap, we set $F_1^{(j)}=0$; otherwise:
\begin{equation}
\label{eq:prf1}
P^{(j)}=\frac{\mathrm{overlap}}{|T(\hat{y})|},\qquad
R^{(j)}=\frac{\mathrm{overlap}}{|T(y^{(j)})|},\qquad
F_1^{(j)}=\frac{2P^{(j)}R^{(j)}}{P^{(j)}+R^{(j)}}.
\end{equation}

When multiple references are available, we take the best-matching one:
\begin{equation}
\label{eq:f1_max_over_refs}
F_1(\hat{y},\mathcal{Y})=\max_{1\le j\le m} F_1^{(j)}.
\end{equation}

For closed-set answers $\{\texttt{yes},\texttt{no}\}$, we use \textbf{exact match (EM)} under normalization:
\begin{equation}
\label{eq:exact_match_yesno}
F_1^{(j)}=
\begin{cases}
1, & \mathrm{norm}(\hat{y})=\mathrm{norm}(y^{(j)}),\\
0, & \text{otherwise}.
\end{cases}
\end{equation}
Finally, the dataset-level token $F_1$ is the mean over samples:
\begin{equation}
\label{eq:f1_dataset}
\mathrm{F1}_{\text{dataset}}=\frac{1}{N}\sum_{i=1}^{N} F_1\bigl(\hat{y}_i,\mathcal{Y}_i\bigr),
\end{equation}
where $\hat{y}_i$ is the prediction for sample $i$ and $\mathcal{Y}_i$ is its reference set.

\paragraph{Inference Cost: Average input tokens (TK).}
We use the average number of input tokens in the \textbf{final answer-generation call} as a proxy for inference cost.
For sample $i$, let $x_i$ be the generator input at the final stage (i.e., the prompt containing the selected evidence and instructions), and let $\mathrm{tok}(\cdot)$ be the generator's tokenizer that maps text to a token sequence.
We define per-sample input tokens and the dataset average as:
\begin{equation}
\label{eq:tk_def}
\mathrm{TK}_i = \bigl|\mathrm{tok}(x_i)\bigr|,\qquad
\mathrm{TK}_{\text{dataset}}=\frac{1}{N}\sum_{i=1}^{N}\mathrm{TK}_i.
\end{equation}
Since Transformer inference cost increases notably with context length (memory and computation), $\mathrm{TK}$ provides a reproducible proxy for throughput/latency when comparing evidence-selection strategies under the same generator.

\paragraph{Cost--quality efficiency: Normalized Token Efficiency (NTE).}
To measure ``better'' and ``cheaper'' jointly, we define the relative token-efficiency improvement of a method $\ell$ over a baseline $b$
(\textsc{Retriever-Only}, e.g., BM25/Contriever).
Let $F1_{\ell}$ and $TK_{\ell}$ denote the dataset-level $\mathrm{F1}_{\text{dataset}}$ and $\mathrm{TK}_{\text{dataset}}$ achieved by method $\ell$, respectively (and analogously $F1_b$, $TK_b$ for the baseline).
We define:
\begin{equation}
\label{eq:nte_def}
\mathrm{NTE}(\ell)
= \frac{F1_{\ell}/TK_{\ell}}{F1_b/TK_b}
= \frac{F1_{\ell}}{F1_b}\bigg/\frac{TK_{\ell}}{TK_b}.
\end{equation}
By definition, $\mathrm{NTE}(b)=1$; $\mathrm{NTE}(\ell)>1$ means method $\ell$ achieves higher answer quality per input token than the baseline.

\paragraph{Retrieval quality: NDCG@k (for analysis, not the optimization target).}
When analyzing the relationship between retrieval ranking quality and end-to-end generation quality (\autoref{sec:relevance-vs-helpfulness}), we additionally report NDCG@k.
Consider the top-$k$ ranked retrieved items $(d_1,\ldots,d_k)$ for a query, where $d_i$ is the item at rank $i$ and $\mathrm{rel}_i\in\{0,1,2,\ldots\}$ is its graded relevance label.
We define discounted cumulative gain (DCG) and its normalized version (NDCG) as:
\begin{equation}
\label{eq:ndcg_def}
\mathrm{DCG@}k=\sum_{i=1}^{k}\frac{2^{\mathrm{rel}_i}-1}{\log_2(i+1)},\qquad
\mathrm{NDCG@}k=\frac{\mathrm{DCG@}k}{\mathrm{IDCG@}k},
\end{equation}
where $\mathrm{IDCG@}k$ is the DCG@k of the ideal ranking obtained by sorting items by descending $\mathrm{rel}_i$.

\paragraph{Selection overhead (probing/scoring calls).}
Some rerankers use the generator as a scorer (e.g., YesNo/QLM), requiring additional model calls over the candidate set;
likewise, IGP requires one unconditional probing rollout and $N$ conditional probing rollouts per query.
Therefore, \textbf{IGP's extra probing cost is of the same order as other LLM-based scoring baselines} (e.g., YesNo/QLM) when applied to the same candidate size $N$.
To keep the cost reporting aligned with \emph{deployment-time context length} under a fixed \TopM interface, we report $\mathrm{TK}$ for the \textbf{final answer-generation stage}; selection overhead is comparable among LLM-based rerankers and can be batched/parallelized in practice (see \autoref{sec:method_practical}).

\subsection{Main Experimental Results (RQ1)}
\label{sec:main_results_bm25}

\autoref{tab:performance-bm25} reports end-to-end performance when using BM25 as the first-stage retriever.
All methods share the same corpus, prompts, generator, decoding settings, candidate size, and the same budget executor \texttt{truncate} (\TopM); we only change the \texttt{rerank}/admission strategy.
Overall, IGP consistently improves the quality--cost trade-off: it reduces input tokens while maintaining or improving answer F1 under the same \TopM interface.

\paragraph{\TopM$=5$: reordering alone is almost ineffective; admission control is the main driver.}
With a wider evidence budget, most reranking methods behave similarly to the retriever-only baseline.
Relevance rerankers (\texttt{+CE}/\texttt{+BGE}) and generator-as-scorer baselines (\texttt{+YesNo}/\texttt{+QLM}) leave the final context nearly unchanged because \texttt{truncate} will still admit five passages, so the token cost remains around the BM25 baseline and F1 changes are marginal.
Likewise, using IG only as a sorting score (\texttt{+IG}) provides little end-to-end gain, indicating that ``utility-aware ordering without filtering'' is insufficient under a wide budget.

In contrast, once threshold-based pruning is enabled, the improvements become substantial.
IGP($T_p{=}0.0$) already reduces TK notably while improving average F1, and a slightly more conservative barrier IGP($T_p{=}0.05$) further pushes the operating point to a markedly better region (higher F1 with much fewer tokens).
This pattern supports our central mechanism: under multi-evidence injection, the bottleneck is often not retrieval recall but \emph{which evidence should be admitted under a limited budget}.
Filtering weak/negative-utility passages reduces redundancy and conflicts that would otherwise spread the generator distribution and destabilize the final answer.

\paragraph{\TopM$=1$: utility-aware admission improves both accuracy and efficiency.}
Under the tightest budget, only one passage can enter the context.
Here, the problem becomes selecting evidence that is not only relevant, but also decisive enough to make generation concentrated on key tokens.
Relevance rerankers improve F1 over BM25 but keep TK almost unchanged, yielding limited efficiency gains.
IGP couples utility-aware ordering with an explicit admission barrier: it improves F1 while simultaneously reducing input tokens by filtering out passages that are relevant yet non-conclusive or ambiguity-inducing.
As a result, IGP achieves consistently stronger quality--cost trade-offs under \TopM$=1$.

\begin{figure}[pos=htbp]
    \centering
    \includegraphics[width=0.9\linewidth]{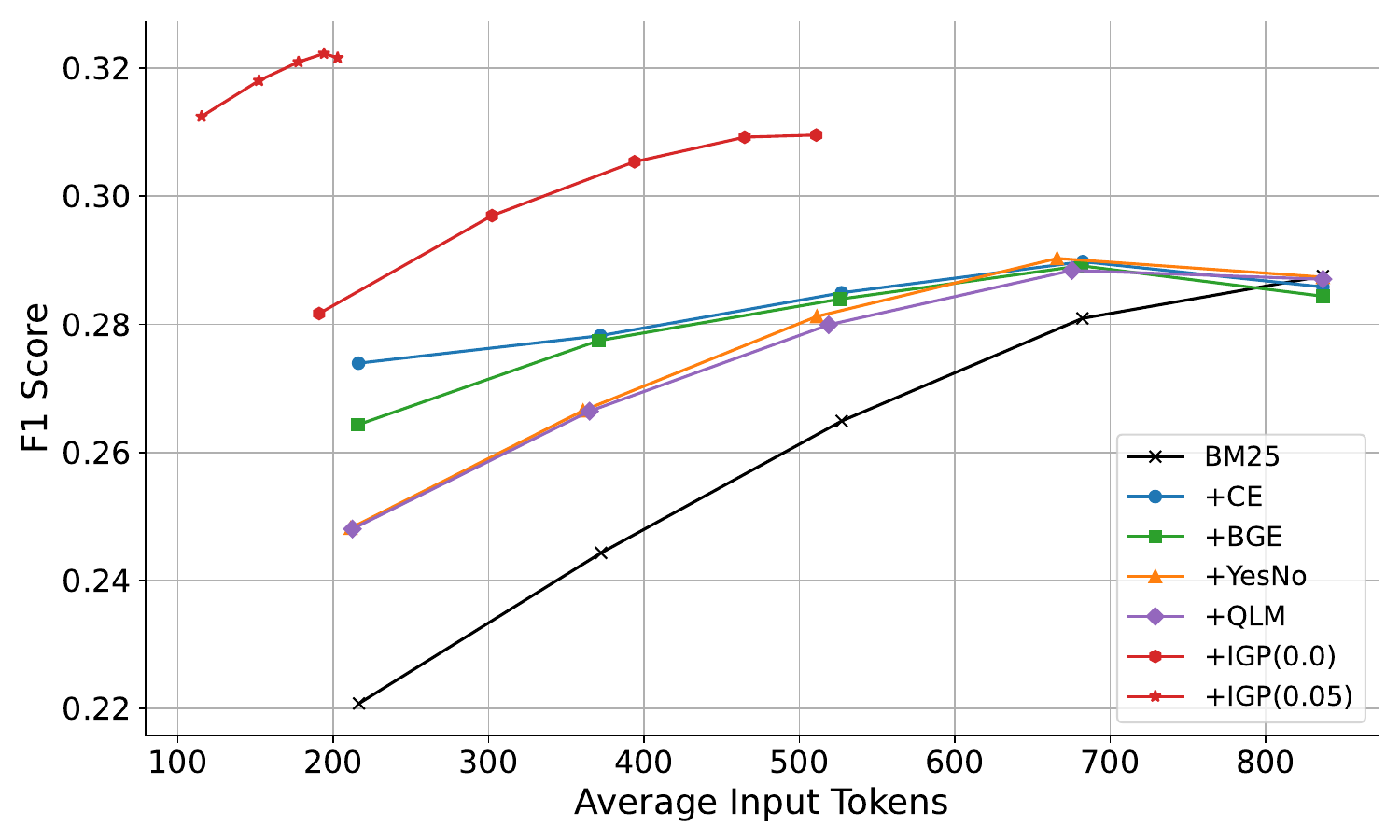}
    \caption{\textbf{Quality--cost Pareto frontiers under BM25 retrieval.}
We plot the cross-dataset average token-level F1 (higher is better) against the average final-stage input tokens TK (lower is better), averaging over \textbf{NQ, TriviaQA, PopQA, SQuAD, and AmbigQA}.
Each curve is obtained by sweeping the evidence budget \TopM; each point corresponds to a feasible deployment operating point under the same \texttt{truncate} budget interface.
Curves closer to the upper-left indicate a better quality--cost trade-off.}
    \label{fig:pareto-bm25}
\end{figure}

\paragraph{Pareto view: IGP strengthens the frontier under BM25.}
\autoref{fig:pareto-bm25} summarizes the budget sweep in a single quality--cost plane.
Across \TopM, conventional rerankers mostly trace curves close to the BM25 baseline, reflecting the limited effect of reordering when most retrieved passages are still admitted.
IGP produces an upper-left shift of the frontier, meaning it reaches comparable (or better) F1 with fewer input tokens, and yields more attractive deployment operating points.

\begin{table}[pos=htbp]
\centering
\setlength{\tabcolsep}{4pt}
\renewcommand{\arraystretch}{1.05}
\resizebox{\linewidth}{!}{

\begin{tabular}{lcccccccccccccccccc}
\toprule
\multirow{2}{*}{Method} & \multicolumn{3}{c}{NQ} & \multicolumn{3}{c}{TriviaQA} & \multicolumn{3}{c}{PopQA} & \multicolumn{3}{c}{SQuAD} & \multicolumn{3}{c}{AmbigQA} & \multicolumn{3}{c}{Avg.} \\
 & F1(↑) & TK(↓) & NTE(↑) & F1(↑) & TK(↓) & NTE(↑) & F1(↑) & TK(↓) & NTE(↑) & F1(↑) & TK(↓) & NTE(↑) & F1(↑) & TK(↓) & NTE(↑) & F1(↑) & TK(↓) & NTE(↑) \\
\midrule
\multicolumn{19}{l}{\textbf{TopM=1}} \\
\midrule
BM25 & .149 & 215.6 & 1.0 & .430 & 224.8 & 1.0 & .169 & 212.5 & 1.0 & .164 & 213.5 & 1.0 & .191 & 216.4 & 1.0 & .221 & 216.6 & 1.0 \\
+CE & .203 & 214.7 & 1.4 & .477 & 224.2 & 1.1 & \underline{\textcolor{secondcolor}{.205}} & 214.7 & 1.2 & \underline{\textcolor{secondcolor}{.219}} & 212.4 & 1.3 & .266 & 215.5 & 1.4 & .274 & 216.3 & 1.2 \\
+BGE & .206 & 213.8 & 1.4 & .469 & 224.0 & 1.1 & .199 & 213.8 & 1.2 & .196 & 211.8 & 1.2 & .252 & 215.5 & 1.3 & .264 & 215.8 & 1.2 \\
+YesNo & .178 & 209.2 & 1.2 & .456 & 220.6 & 1.1 & .186 & 207.8 & 1.1 & .187 & 208.0 & 1.2 & .233 & 210.6 & 1.3 & .248 & 211.2 & 1.2 \\
+QLM & .176 & 208.5 & 1.2 & .464 & 221.6 & 1.1 & .186 & 210.2 & 1.1 & .210 & 211.2 & 1.3 & .204 & 210.4 & 1.1 & .248 & 212.4 & 1.1 \\
+IG & .185 & 214.5 & 1.2 & .457 & 224.8 & 1.1 & .179 & 212.5 & 1.1 & .188 & 213.7 & 1.1 & .237 & 215.9 & 1.2 & .249 & 216.3 & 1.1 \\
+IGP(0.0) & \underline{\textcolor{secondcolor}{.221}} & \underline{\textcolor{secondcolor}{182.5}} & \underline{\textcolor{secondcolor}{1.8}} & \underline{\textcolor{secondcolor}{.494}} & \underline{\textcolor{secondcolor}{196.0}} & \underline{\textcolor{secondcolor}{1.3}} & .196 & \underline{\textcolor{secondcolor}{197.9}} & \underline{\textcolor{secondcolor}{1.2}} & .209 & \underline{\textcolor{secondcolor}{194.0}} & \underline{\textcolor{secondcolor}{1.4}} & \underline{\textcolor{secondcolor}{.289}} & \underline{\textcolor{secondcolor}{184.2}} & \underline{\textcolor{secondcolor}{1.8}} & \underline{\textcolor{secondcolor}{.282}} & \underline{\textcolor{secondcolor}{190.9}} & \underline{\textcolor{secondcolor}{1.4}} \\
+IGP(0.05) & \textbf{\textcolor{bestcolor}{.253}} & \textbf{\textcolor{bestcolor}{103.8}} & \textbf{\textcolor{bestcolor}{3.5}} & \textbf{\textcolor{bestcolor}{.513}} & \textbf{\textcolor{bestcolor}{102.9}} & \textbf{\textcolor{bestcolor}{2.6}} & \textbf{\textcolor{bestcolor}{.228}} & \textbf{\textcolor{bestcolor}{137.0}} & \textbf{\textcolor{bestcolor}{2.1}} & \textbf{\textcolor{bestcolor}{.235}} & \textbf{\textcolor{bestcolor}{123.2}} & \textbf{\textcolor{bestcolor}{2.5}} & \textbf{\textcolor{bestcolor}{.334}} & \textbf{\textcolor{bestcolor}{109.5}} & \textbf{\textcolor{bestcolor}{3.4}} & \textbf{\textcolor{bestcolor}{.312}} & \textbf{\textcolor{bestcolor}{115.3}} & \textbf{\textcolor{bestcolor}{2.7}} \\
\midrule
\multicolumn{19}{l}{\textbf{TopM=5}} \\
\midrule
BM25 & .214 & 835.2 & 1.0 & .504 & 853.5 & 1.0 & .211 & 833.4 & 1.0 & .227 & 822.8 & 1.0 & .282 & 839.5 & 1.0 & .288 & 836.9 & 1.0 \\
+CE & .212 & 835.2 & 1.0 & .506 & 853.5 & 1.0 & .209 & 833.4 & 1.0 & .226 & 822.8 & 1.0 & .276 & 839.5 & 1.0 & .286 & 836.9 & 1.0 \\
+BGE & .211 & 835.2 & 1.0 & .505 & 853.5 & 1.0 & .210 & 833.4 & 1.0 & .225 & 822.8 & 1.0 & .271 & 839.5 & 1.0 & .284 & 836.9 & 1.0 \\
+YesNo & .214 & 835.2 & 1.0 & .506 & 853.5 & 1.0 & .211 & 833.4 & 1.0 & .225 & 822.8 & 1.0 & .280 & 839.5 & 1.0 & .287 & 836.9 & 1.0 \\
+QLM & .212 & 835.2 & 1.0 & .506 & 853.5 & 1.0 & .210 & 833.4 & 1.0 & .228 & 822.8 & 1.0 & .279 & 839.5 & 1.0 & .287 & 836.9 & 1.0 \\
+IG & .214 & 835.2 & 1.0 & .505 & 853.5 & 1.0 & .208 & 833.4 & 1.0 & .227 & 822.8 & 1.0 & .284 & 839.5 & 1.0 & .288 & 836.9 & 1.0 \\
+IGP(0.0) & \underline{\textcolor{secondcolor}{.242}} & \underline{\textcolor{secondcolor}{463.5}} & \underline{\textcolor{secondcolor}{2.0}} & \textbf{\textcolor{bestcolor}{.527}} & \underline{\textcolor{secondcolor}{486.8}} & \underline{\textcolor{secondcolor}{1.8}} & \underline{\textcolor{secondcolor}{.219}} & \underline{\textcolor{secondcolor}{585.3}} & \underline{\textcolor{secondcolor}{1.5}} & \underline{\textcolor{secondcolor}{.238}} & \underline{\textcolor{secondcolor}{537.4}} & \underline{\textcolor{secondcolor}{1.6}} & \underline{\textcolor{secondcolor}{.321}} & \underline{\textcolor{secondcolor}{481.1}} & \underline{\textcolor{secondcolor}{2.0}} & \underline{\textcolor{secondcolor}{.310}} & \underline{\textcolor{secondcolor}{510.8}} & \underline{\textcolor{secondcolor}{1.8}} \\
+IGP(0.05) & \textbf{\textcolor{bestcolor}{.257}} & \textbf{\textcolor{bestcolor}{168.2}} & \textbf{\textcolor{bestcolor}{6.0}} & \underline{\textcolor{secondcolor}{.520}} & \textbf{\textcolor{bestcolor}{155.3}} & \textbf{\textcolor{bestcolor}{5.7}} & \textbf{\textcolor{bestcolor}{.238}} & \textbf{\textcolor{bestcolor}{281.6}} & \textbf{\textcolor{bestcolor}{3.3}} & \textbf{\textcolor{bestcolor}{.248}} & \textbf{\textcolor{bestcolor}{225.1}} & \textbf{\textcolor{bestcolor}{4.0}} & \textbf{\textcolor{bestcolor}{.345}} & \textbf{\textcolor{bestcolor}{184.0}} & \textbf{\textcolor{bestcolor}{5.6}} & \textbf{\textcolor{bestcolor}{.322}} & \textbf{\textcolor{bestcolor}{202.9}} & \textbf{\textcolor{bestcolor}{4.6}} \\
\bottomrule
\end{tabular}

} 
\caption{\textbf{End-to-end RAG results with BM25 retrieval.}
We compare reranking/admission strategies under a fixed pipeline \texttt{retrieve$\rightarrow$rerank$\rightarrow$truncate}, where only the \texttt{rerank} stage is replaced (IGP) and the budget executor \texttt{truncate} remains unchanged.
We report token-level answer F1 (higher is better), average final-stage input tokens TK (lower is better), and normalized token efficiency NTE (higher is better; relative to the retriever-only baseline under the same \TopM).
Results are shown for two evidence budgets (\TopM$=1$ and \TopM$=5$) on five QA benchmarks and their average.
IGP($T_p$) denotes Information Gain Pruning with pruning threshold $T_p$ (admission barrier).}
\label{tab:performance-bm25}
\end{table}

\subsection{Ablation Study (RQ2)}
\label{sec:ablation_results}

This subsection answers RQ2: how sensitive IGP is to the pruning threshold $T_p$, and how uncertainty-estimation hyperparameters (\TopK, $MT$) affect stability and end-to-end performance.

\subsubsection{Sensitivity to the IGP Threshold $T_p$ (RQ2)}
\label{sec:ig_threshold_sensitivity}

\begin{figure}[pos=htbp]
\centering
\includegraphics[width=0.9\linewidth]{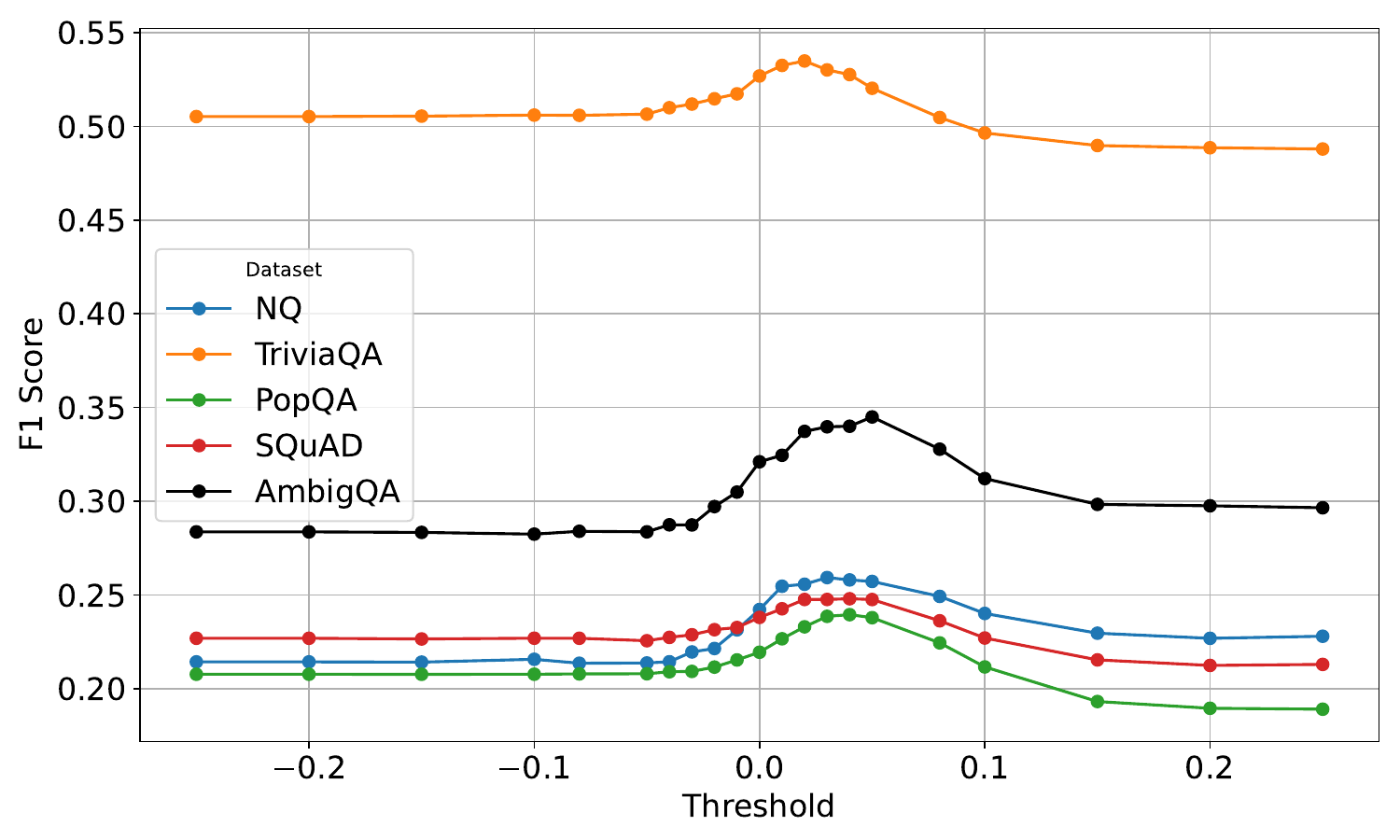}
\caption{\textbf{Sensitivity of IGP to the pruning threshold $T_p$ under \TopM$=1$ (token-level F1).}}
\label{fig:ig-thres-f1}
\end{figure}

\paragraph{A stable ``sweet spot'': too low introduces noise, too high reduces coverage.}
Under a tight evidence budget (\TopM$=1$), $T_p$ is essentially the admission barrier for the single evidence slot:
a candidate passage is admitted only if
$IG(d,q;\phi,K)$ (Eq.~\eqref{eq:ig_def_method}) is not smaller than $T_p$.
Thus, $T_p$ directly controls the trade-off between \emph{denoising strength} and \emph{coverage}.

\paragraph{A clear unimodal trend across datasets, with the peak near small positive values.}
We fix all other settings as in the main experiments (\autoref{sec:exp_setup}) and scan $T_p$ under \TopM$=1$.
As shown in \autoref{fig:ig-thres-f1}, when $T_p$ increases from negative/near-zero values to a \emph{small positive} range, end-to-end F1 rises and reaches a peak; when $T_p$ increases further, F1 gradually drops.
This unimodal pattern appears consistently on NQ, TriviaQA, PopQA, SQuAD, and AmbigQA.

\paragraph{Why: low thresholds lose to ``noisy/conflicting occupancy,'' high thresholds lose to ``missing evidence.''}
When $T_p$ is low, weak-gain or negative-gain passages ($IG\le 0$) are more likely to pass admission control, occupying the only evidence slot with more discussion-like or multi-claim content.
Such evidence can introduce competing hypotheses or conditional constraints, making the generation distribution less concentrated on the correct answer tokens and lowering F1.
When $T_p$ is too high, some passages that are necessary or complementary (but with smaller single-passage gain) may also be filtered out, making the context miss key details and again lowering F1.
Therefore, the optimal region is usually in the middle: filter weak/negative-utility evidence while not sacrificing too much coverage.

\paragraph{The sweet spot is somewhat transferable, but the absolute scale of $T_p$ is not universal.}
\autoref{fig:ig-thres-f1} shows that peaks cluster in a similar range (often around $T_p\approx 0.05$ in our setting), suggesting that large-scale tuning per dataset is usually unnecessary.
However, the absolute value of $T_p$ depends on the distribution of $IG$, which can change with the generator family/scale, tokenization and decoding details, and the question distribution.
Thus, when switching generators or tasks, it is better to scan within a \emph{narrow range} to re-locate the sweet spot, rather than assuming a fixed constant transfers directly.

\subsubsection{Sensitivity of Uncertainty Estimation: \TopK and MaxTokens ($MT$) (RQ2)}
\label{sec:topk_mt_sensitivity}

\begin{figure}[pos=htbp]
    \centering

    \begin{subfigure}[t]{0.9\linewidth}
        \centering
        \includegraphics[width=\linewidth]{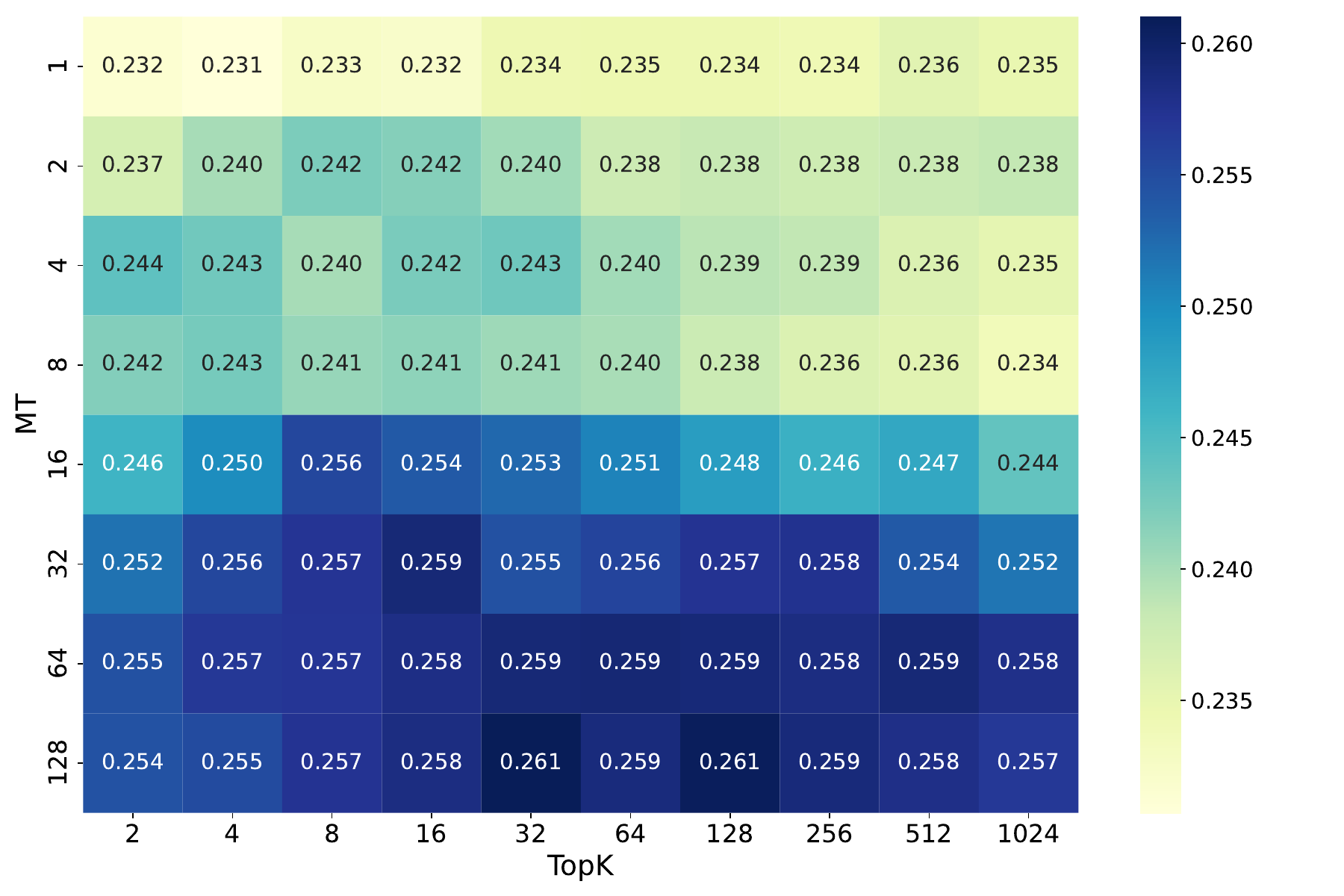}
        \caption{\TopK--$MT$ heatmap.}
        \label{fig:heatmap-of-topk-maxtk}
    \end{subfigure}

    \vspace{0.6em}

    \begin{subfigure}[t]{0.49\linewidth}
        \centering
        \includegraphics[width=\linewidth]{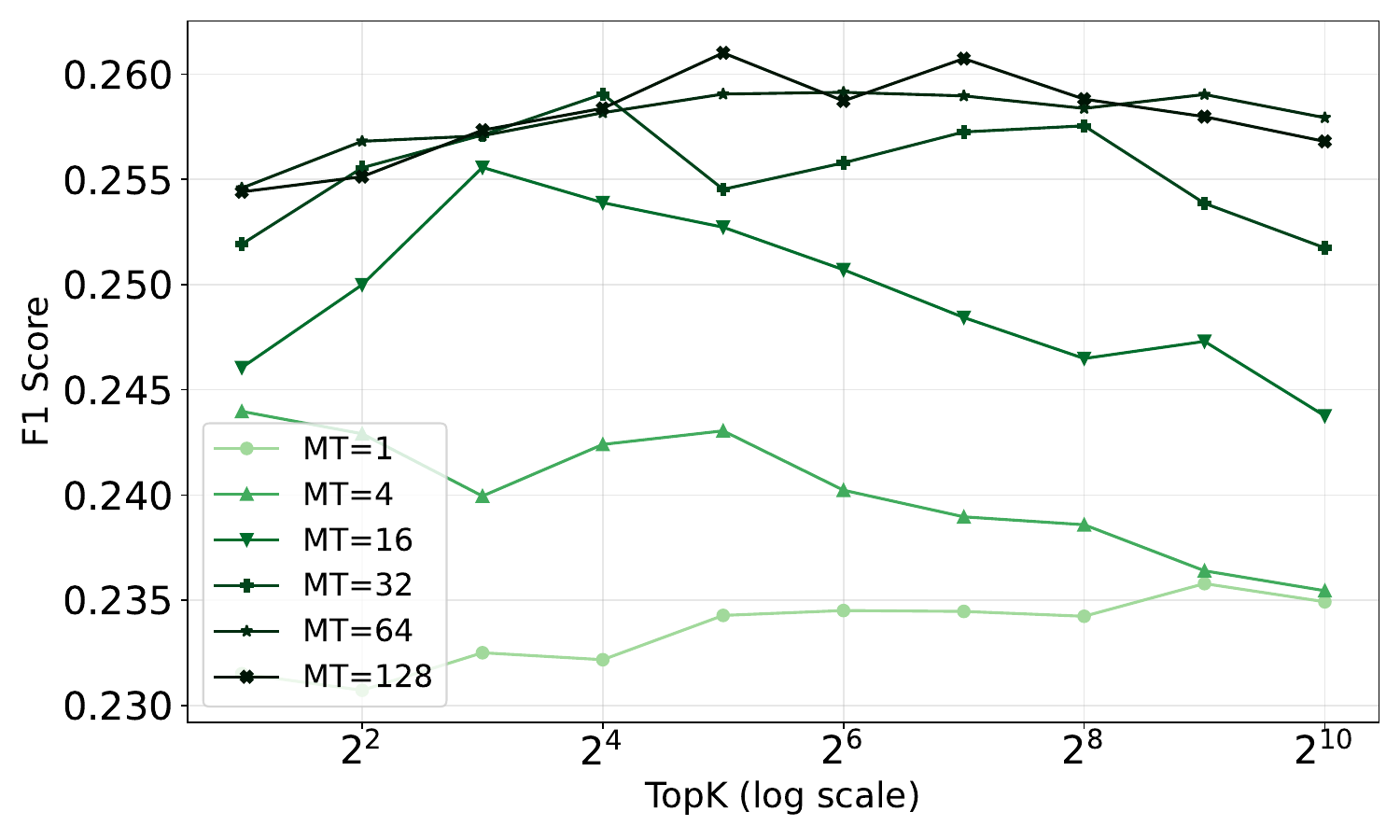}
        \caption{\TopK--F1.}
        \label{fig:topk-f1}
    \end{subfigure}
    \hfill
    \begin{subfigure}[t]{0.49\linewidth}
        \centering
        \includegraphics[width=\linewidth]{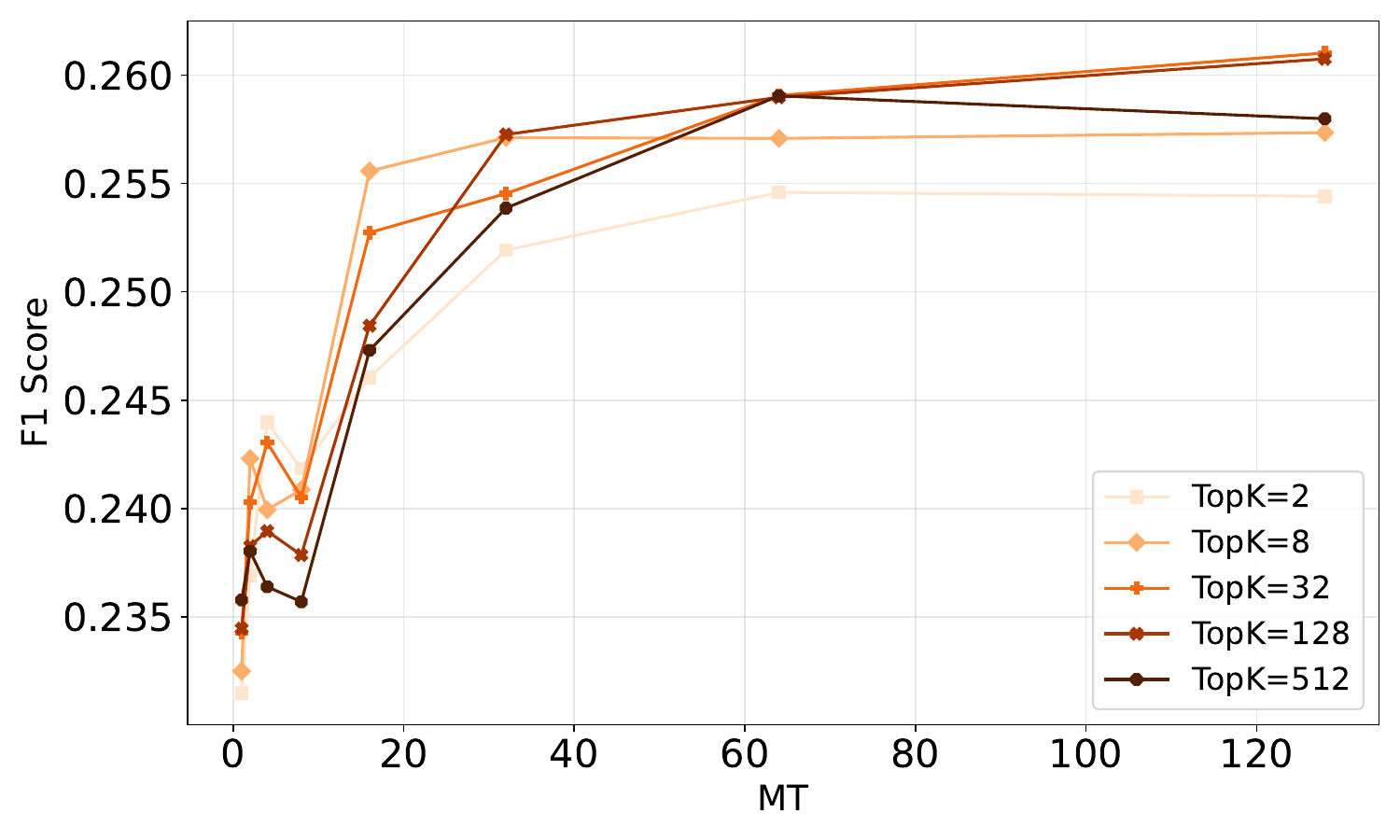}
        \caption{$MT$--F1.}
        \label{fig:maxtk-f1}
    \end{subfigure}

    \caption{\textbf{Sensitivity to \TopK and maximum rollout length $MT$ (end-to-end F1).}
    (a) A heatmap of F1 over a 2D grid of $(K,MT)$;
    (b) the marginal curve of \TopK with fixed $MT$;
    (c) the marginal curve of $MT$ with fixed \TopK.
    Here $K$ is the truncation size for \TopK renormalization (Eq.~\eqref{eq:topk_softmax_method}), and $MT$ is the maximum rollout length with the effective-length rule in Eq.~\eqref{eq:mt_trunc_rule_method}.
    Overall, $MT$ plays a more dominant role, while \TopK becomes robust once it reaches moderate values.}
    \label{fig:topk-maxtk-combined}
\end{figure}

\paragraph{$MT$ matters more than \TopK, and both show clear plateaus at moderate values.}
The IGP signal depends on uncertainty estimation: under greedy rollout (temperature $0$), we construct a \TopK-truncated and renormalized distribution (Eq.~\eqref{eq:topk_softmax_method}), compute token-level entropy and normalized uncertainty (Eq.~\eqref{eq:topk_token_entropy_method}--\eqref{eq:topk_normalized_uncertainty_method}), average over time to obtain $\widehat{NU}$ (Eq.~\eqref{eq:nu_hat_uncond_method}), and define information gain as uncertainty reduction (Eq.~\eqref{eq:ig_def_method}).
Therefore, \TopK and $MT$ affect the variance and resolution of $\widehat{NU}$, which in turn affects IGP reranking and threshold-based admission.
We systematically scan $(K,MT)$ while keeping other settings fixed (\autoref{sec:exp_setup}), and evaluate end-to-end F1 (see \autoref{fig:topk-maxtk-combined}).

\paragraph{Increasing $MT$ yields larger and more stable gains, then gradually plateaus.}
When $MT$ increases from very small to moderate values, end-to-end F1 improves noticeably; further increases bring smaller gains (\autoref{fig:heatmap-of-topk-maxtk} and \autoref{fig:maxtk-f1}).
This matches the form of $\widehat{NU}$: it is a time average of token-level uncertainty $\tilde{u}_t$ (Eq.~\eqref{eq:nu_hat_uncond_method}), and $IG$ is the difference between two uncertainty estimates (Eq.~\eqref{eq:ig_def_method}).
With too-short rollouts, $\widehat{NU}$ can be dominated by early-step fluctuations, increasing estimation noise and causing ranking flips among candidates.
A larger $MT$ effectively averages over a longer trajectory, reducing variance and stabilizing relative ordering.
Once $MT$ reaches the plateau, the marginal benefit becomes small, indicating that the relative ranking signal needed by IGP is already stable.

\paragraph{\TopK mainly affects resolution: avoid very small $K$, and moderate $K$ is often enough.}
From \autoref{fig:heatmap-of-topk-maxtk} and \autoref{fig:topk-f1}, very small $K$ tends to hurt F1, while performance becomes flat once $K$ reaches a moderate scale; further increasing $K$ brings limited gains and may slightly drop in some settings.
This is consistent with \TopK truncation: if $K$ is too small, probability mass on competing tokens is discarded, making $\tilde{H}_t$ (Eq.~\eqref{eq:topk_token_entropy_method}) saturate or distort and reducing the resolution of $\widehat{NU}$.
Once $K$ covers major competing tokens, the relative $IG$ comparisons become stable; increasing $K$ then mainly increases the cost of returning/processing \TopK log-probabilities, especially in black-box settings.

\paragraph{Prioritize $MT$ before tuning $K$: large $K$ cannot fix instability from short rollouts.}
The heatmap shows a clear high-performance band: when $MT$ is small, increasing $K$ cannot overcome the performance limit; when $MT$ is moderate or larger, many moderate $K$ values fall into a similar high-performance plateau (\autoref{fig:heatmap-of-topk-maxtk}).
This suggests that $MT$ acts as a stability requirement, while $K$ is more like a resolution adjustment: first ensure the rollout is long enough to stabilize $IG$ ranking, then choose a moderate $K$ within interface cost constraints.

\subsection{More Experimental Results and Discussion (RQ3--RQ5)}
\label{sec:more_results}

This subsection provides additional analysis and robustness results.
We first analyze the relevance--utility mismatch (RQ3), then evaluate robustness to the first-stage retriever (RQ4) and to generator family/scale (RQ5).

\subsubsection{Relevance Metrics vs.\ End-to-End Generation Quality (RQ3)}
\label{sec:relevance-vs-helpfulness}

Offline retrieval metrics (e.g., NDCG@k) are widely used to evaluate evidence ranking, but the end goal in RAG is end-to-end answer quality (token-level F1; \autoref{sec:metrics}).
Here we provide controlled evidence that \textbf{relevance is not a reliable proxy for evidence utility to generation}, and the mismatch becomes more pronounced under multi-evidence budgets.

\paragraph{Setting (controlled comparison).}
We evaluate on the \textbf{NQ test set} and compute retrieval relevance using \textbf{BEIR-NQ} annotations, while generation quality is measured by token-level \textbf{F1} against NQ short answers.
We take the intersection of instances that have both relevance labels and gold answers (2,724 questions).
We use \textbf{BM25} retrieval with top-$n{=}5$ candidates, and vary \emph{only} the evidence ordering/admission strategy (rerank/prune), keeping the generator (\textbf{Qwen2.5-7B-Instruct}), prompts, and decoding fixed.
We report results under two budgets: \TopM$=1$ and \TopM$=5$ (\autoref{tab:ndcg-vs-f1}), and compute Spearman correlation between NDCG and F1 (see \autoref{fig:retrieval-generation-corr}).

\begin{table}[pos=htbp]
\centering
\setlength{\tabcolsep}{4pt}
\renewcommand{\arraystretch}{1.05}

\begin{tabular}{lcccc}
\toprule
\multirow{2}{*}{Method} & \multicolumn{2}{c}{TopM=1} & \multicolumn{2}{c}{TopM=5} \\
 & NDCG & F1 & NDCG & F1 \\
\midrule
BM25 & 0.0519 & 0.1761 & 0.0816 & 0.2298 \\
BGE & 0.0904 & 0.2172 & 0.0996 & 0.2312 \\
CE & 0.0918 & 0.2208 & 0.1008 & 0.2295 \\
YesNo & 0.0542 & 0.1910 & 0.0826 & 0.2314 \\
QLM & 0.0591 & 0.1898 & 0.0845 & 0.2329 \\
IGP(0.0) & 0.0716 & 0.2400 & 0.0841 & 0.2535 \\
IGP(0.05) & 0.0417 & 0.2610 & 0.0464 & 0.2675 \\
\bottomrule
\end{tabular}

\caption{\textbf{Relevance--utility mismatch in budgeted RAG.}
Under a controlled pipeline on NQ (2,724 instances), higher NDCG does not consistently yield higher end-to-end F1, and the mismatch can worsen when multiple passages are injected.}
\label{tab:ndcg-vs-f1}
\end{table}

\paragraph{Observation 1: tight budget (\TopM$=1$) shows weak rank agreement.}
With only one passage injected, \textbf{higher NDCG@1 does not consistently imply higher F1}.
A passage can be topically relevant yet \emph{non-decisive} (e.g., containing multiple conditions or parallel claims), which can keep the generator uncertain and yield unstable answers, leading to weak Spearman rank correlation (\autoref{fig:retrieval-generation-corr}).

\paragraph{Observation 2: wide budget (\TopM$=5$) can produce negative correlation.}
When injecting multiple passages, \textbf{the correlation between NDCG@5 and F1 can become negative} (\autoref{fig:retrieval-generation-corr}).
This reflects an amplification effect: even highly relevant passages may be redundant or mildly inconsistent in scope/granularity/details, and \emph{their combination} can spread the generator distribution and reduce answer stability.

\paragraph{Direct evidence of mismatch.}
\autoref{tab:ndcg-vs-f1} shows a representative pattern: \textbf{IGP(0.05) attains lower NDCG but higher F1} under both \TopM$=1$ and \TopM$=5$.
This indicates that optimizing relevance alone can over-admit relevant-but-ambiguity-inducing evidence and miss passages that are more decisive for generation under a fixed budget.

\paragraph{Implication.}
These results explain why reranking-by-relevance (or reordering alone) often yields limited end-to-end gains under \TopM truncation: many retrieved passages are still admitted, so redundancy/conflicts remain.
In contrast, \textbf{utility-aware admission control} (IGP thresholding) can more directly improve end-to-end quality by filtering weak/negative-utility evidence before injection.

\subsubsection{Retriever Robustness via Pareto Frontiers (RQ4)}
\label{sec:retriever_effect}

\begin{figure}[pos=htbp]
    \centering
    \includegraphics[width=0.9\linewidth]{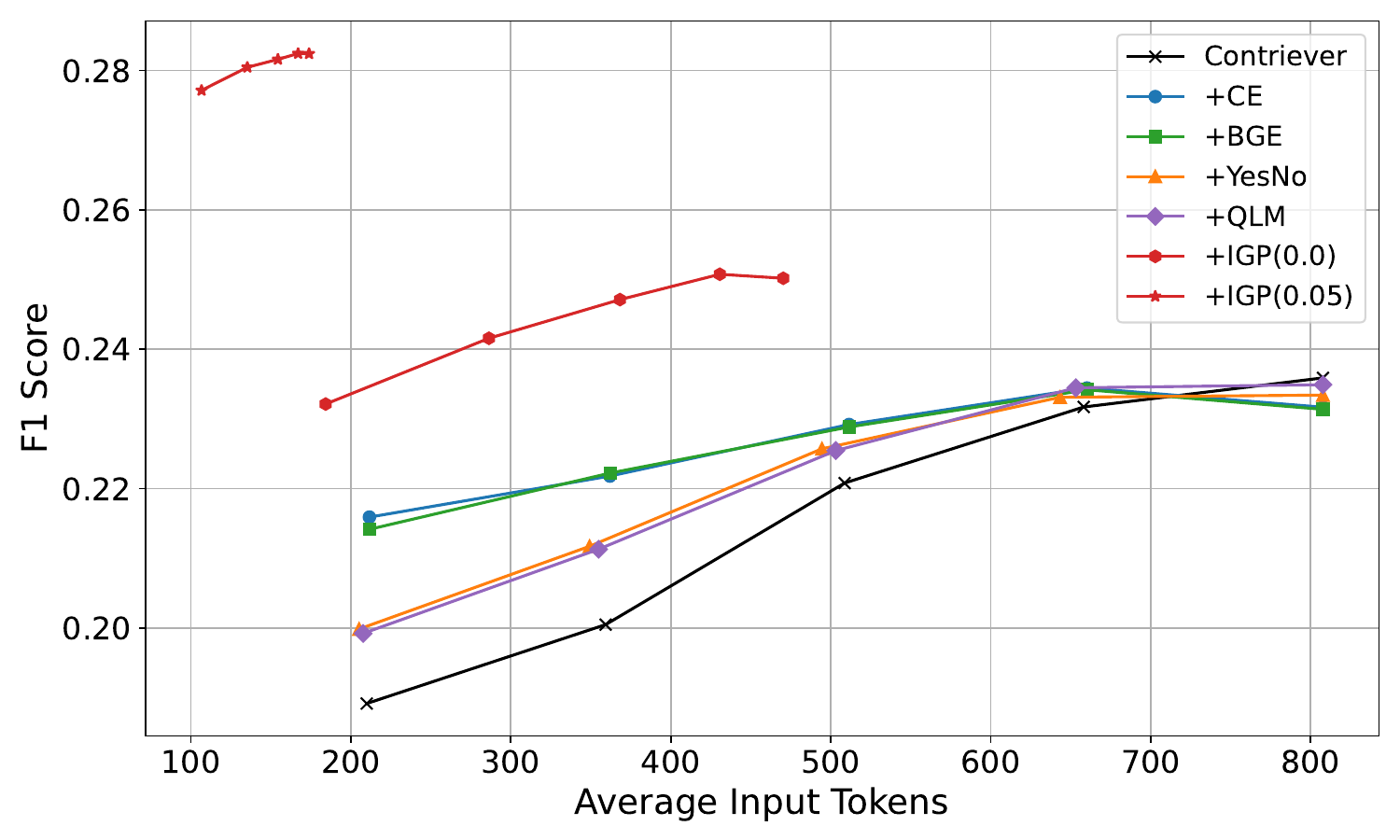}
    \caption{\textbf{Quality--cost Pareto frontiers under Contriever retrieval.}
Same protocol as \autoref{fig:pareto-bm25}, but with Contriever as the first-stage retriever.
We sweep \TopM to obtain deployment operating points and plot average F1 versus average TK over NQ, TriviaQA, PopQA, SQuAD, and AmbigQA.
Across retrievers, a consistent upper-left shift indicates retriever-robust improvements in the quality--cost trade-off.}
    \label{fig:pareto-contriever}
\end{figure}

We test whether IGP depends on a particular retriever by switching the first-stage retriever from sparse BM25 to the unsupervised dense retriever Contriever, while keeping the corpus, candidate size, prompts, decoding protocol, and the \TopM budget interface fixed.
Only the \texttt{rerank}/admission strategy changes across methods.
Quantitative results are reported in \autoref{tab:performance-contriever}, and the budget-sweep Pareto frontier is shown in \autoref{fig:pareto-contriever}.

\paragraph{Consistent frontier shift across retrievers.}
Comparing \autoref{fig:pareto-bm25} (BM25) and \autoref{fig:pareto-contriever} (Contriever), we observe the same qualitative behavior:
conventional rerankers (CE/BGE) and relevance-like zero-shot signals (YesNo/QLM) remain close to the retriever-only frontier, whereas IGP yields operating points closer to the upper-left region.
This indicates that the gains of IGP are not tied to a specific retrieval paradigm; rather, they come from a generator-aligned admission signal that suppresses weak/conflicting evidence before context injection.

\paragraph{Table evidence: pruning dominates under wide budgets, and remains beneficial under tight budgets.}
\autoref{tab:performance-contriever} mirrors the main observations under BM25 (\autoref{tab:performance-bm25}).
Under \TopM$=5$, reordering-based reranking changes little because five passages are still admitted; IGP’s pruning reduces TK substantially and often improves (or maintains) F1.
Under \TopM$=1$, where the single evidence slot is the dominant bottleneck, IGP improves F1 while lowering TK, suggesting that uncertainty-reduction-based utility better matches what the generator needs to converge.

\begin{table}[pos=htbp]
\centering
\setlength{\tabcolsep}{4pt}
\renewcommand{\arraystretch}{1.05}
\resizebox{\linewidth}{!}{

\begin{tabular}{lcccccccccccccccccc}
\toprule
\multirow{2}{*}{Method} & \multicolumn{3}{c}{NQ} & \multicolumn{3}{c}{TriviaQA} & \multicolumn{3}{c}{PopQA} & \multicolumn{3}{c}{SQuAD} & \multicolumn{3}{c}{AmbigQA} & \multicolumn{3}{c}{Avg.} \\
 & F1(↑) & TK(↓) & NTE(↑) & F1(↑) & TK(↓) & NTE(↑) & F1(↑) & TK(↓) & NTE(↑) & F1(↑) & TK(↓) & NTE(↑) & F1(↑) & TK(↓) & NTE(↑) & F1(↑) & TK(↓) & NTE(↑) \\
\midrule
\multicolumn{19}{l}{\textbf{TopM=1}} \\
\midrule
Contriever & .161 & 208.7 & 1.0 & .357 & 218.3 & 1.0 & .102 & 208.2 & 1.0 & .141 & 206.9 & 1.0 & .185 & 207.7 & 1.0 & .189 & 210.0 & 1.0 \\
+CE & .185 & 210.4 & 1.1 & .384 & 220.0 & 1.1 & .114 & 209.0 & 1.1 & .174 & 208.6 & 1.2 & .222 & 210.2 & 1.2 & .216 & 211.6 & 1.1 \\
+BGE & .187 & 210.2 & 1.1 & .383 & 219.8 & 1.1 & .114 & 209.9 & 1.1 & .169 & 208.5 & 1.2 & .219 & 210.3 & 1.2 & .214 & 211.7 & 1.1 \\
+YesNo & .164 & 204.2 & 1.0 & .373 & 213.5 & 1.1 & .105 & 201.7 & 1.1 & .160 & 203.0 & 1.2 & .197 & 203.4 & 1.1 & .200 & 205.2 & 1.1 \\
+QLM & .168 & 204.2 & 1.1 & .378 & 216.1 & 1.1 & .106 & 207.4 & 1.0 & .165 & 206.5 & 1.2 & .180 & 204.1 & 1.0 & .199 & 207.7 & 1.1 \\
+IG & .179 & 209.4 & 1.1 & .370 & 218.8 & 1.0 & .106 & 209.1 & 1.0 & .159 & 208.1 & 1.1 & .203 & 208.7 & 1.1 & .203 & 210.8 & 1.1 \\
+IGP(0.0) & \underline{\textcolor{secondcolor}{.209}} & \underline{\textcolor{secondcolor}{177.6}} & \underline{\textcolor{secondcolor}{1.5}} & \underline{\textcolor{secondcolor}{.408}} & \underline{\textcolor{secondcolor}{187.4}} & \underline{\textcolor{secondcolor}{1.3}} & \underline{\textcolor{secondcolor}{.127}} & \underline{\textcolor{secondcolor}{190.7}} & \underline{\textcolor{secondcolor}{1.4}} & \underline{\textcolor{secondcolor}{.175}} & \underline{\textcolor{secondcolor}{189.6}} & \underline{\textcolor{secondcolor}{1.4}} & \underline{\textcolor{secondcolor}{.242}} & \underline{\textcolor{secondcolor}{176.1}} & \underline{\textcolor{secondcolor}{1.5}} & \underline{\textcolor{secondcolor}{.232}} & \underline{\textcolor{secondcolor}{184.3}} & \underline{\textcolor{secondcolor}{1.4}} \\
+IGP(0.05) & \textbf{\textcolor{bestcolor}{.231}} & \textbf{\textcolor{bestcolor}{99.4}} & \textbf{\textcolor{bestcolor}{3.0}} & \textbf{\textcolor{bestcolor}{.481}} & \textbf{\textcolor{bestcolor}{95.7}} & \textbf{\textcolor{bestcolor}{3.1}} & \textbf{\textcolor{bestcolor}{.168}} & \textbf{\textcolor{bestcolor}{123.0}} & \textbf{\textcolor{bestcolor}{2.8}} & \textbf{\textcolor{bestcolor}{.209}} & \textbf{\textcolor{bestcolor}{116.8}} & \textbf{\textcolor{bestcolor}{2.6}} & \textbf{\textcolor{bestcolor}{.297}} & \textbf{\textcolor{bestcolor}{98.4}} & \textbf{\textcolor{bestcolor}{3.4}} & \textbf{\textcolor{bestcolor}{.277}} & \textbf{\textcolor{bestcolor}{106.7}} & \textbf{\textcolor{bestcolor}{2.9}} \\
\midrule
\multicolumn{19}{l}{\textbf{TopM=5}} \\
\midrule
Contriever & .205 & 806.0 & 1.0 & .420 & 823.0 & 1.0 & .120 & 813.6 & 1.0 & .187 & 793.6 & 1.0 & .247 & 802.4 & 1.0 & .236 & 807.7 & 1.0 \\
+CE & .197 & 806.0 & 1.0 & .416 & 823.0 & 1.0 & .121 & 813.6 & 1.0 & .185 & 793.6 & 1.0 & .240 & 802.4 & 1.0 & .232 & 807.7 & 1.0 \\
+BGE & .197 & 806.0 & 1.0 & .415 & 823.0 & 1.0 & .120 & 813.6 & 1.0 & .188 & 793.6 & 1.0 & .237 & 802.4 & 1.0 & .231 & 807.7 & 1.0 \\
+YesNo & .201 & 806.0 & 1.0 & .419 & 823.0 & 1.0 & .120 & 813.6 & 1.0 & .185 & 793.6 & 1.0 & .243 & 802.4 & 1.0 & .233 & 807.7 & 1.0 \\
+QLM & .200 & 806.0 & 1.0 & .419 & 823.0 & 1.0 & .120 & 813.6 & 1.0 & .187 & 793.6 & 1.0 & .248 & 802.4 & 1.0 & .235 & 807.7 & 1.0 \\
+IG & .200 & 806.0 & 1.0 & .421 & 823.0 & 1.0 & .120 & 813.6 & 1.0 & .188 & 793.6 & 1.0 & .240 & 802.4 & 1.0 & .234 & 807.7 & 1.0 \\
+IGP(0.0) & \underline{\textcolor{secondcolor}{.221}} & \underline{\textcolor{secondcolor}{432.7}} & \underline{\textcolor{secondcolor}{2.0}} & \underline{\textcolor{secondcolor}{.438}} & \underline{\textcolor{secondcolor}{444.1}} & \underline{\textcolor{secondcolor}{1.9}} & \underline{\textcolor{secondcolor}{.136}} & \underline{\textcolor{secondcolor}{541.4}} & \underline{\textcolor{secondcolor}{1.7}} & \underline{\textcolor{secondcolor}{.193}} & \underline{\textcolor{secondcolor}{508.3}} & \underline{\textcolor{secondcolor}{1.6}} & \underline{\textcolor{secondcolor}{.263}} & \underline{\textcolor{secondcolor}{425.1}} & \underline{\textcolor{secondcolor}{2.0}} & \underline{\textcolor{secondcolor}{.250}} & \underline{\textcolor{secondcolor}{470.3}} & \underline{\textcolor{secondcolor}{1.8}} \\
+IGP(0.05) & \textbf{\textcolor{bestcolor}{.236}} & \textbf{\textcolor{bestcolor}{150.4}} & \textbf{\textcolor{bestcolor}{6.2}} & \textbf{\textcolor{bestcolor}{.487}} & \textbf{\textcolor{bestcolor}{135.2}} & \textbf{\textcolor{bestcolor}{7.1}} & \textbf{\textcolor{bestcolor}{.171}} & \textbf{\textcolor{bestcolor}{238.7}} & \textbf{\textcolor{bestcolor}{4.9}} & \textbf{\textcolor{bestcolor}{.216}} & \textbf{\textcolor{bestcolor}{199.8}} & \textbf{\textcolor{bestcolor}{4.6}} & \textbf{\textcolor{bestcolor}{.302}} & \textbf{\textcolor{bestcolor}{145.3}} & \textbf{\textcolor{bestcolor}{6.8}} & \textbf{\textcolor{bestcolor}{.282}} & \textbf{\textcolor{bestcolor}{173.9}} & \textbf{\textcolor{bestcolor}{5.6}} \\
\bottomrule
\end{tabular}

} 
\caption{\textbf{End-to-end RAG results with Contriever retrieval.}
We repeat the same comparison as \autoref{tab:performance-bm25} but switch the first-stage retriever from BM25 to Contriever, while keeping the corpus, prompts, decoding protocol, and the \TopM budget interface fixed.
Only the \texttt{rerank}/admission strategy changes across methods.
We report token-level F1, average final-stage input tokens TK, and NTE (relative to the retriever-only baseline under the same \TopM).}
\label{tab:performance-contriever}
\end{table}

\subsubsection{Generator Family and Scale Robustness (RQ5)}
\label{sec:model_scale}

\begin{figure}[pos=htbp]
\centering
\includegraphics[width=0.9\linewidth]{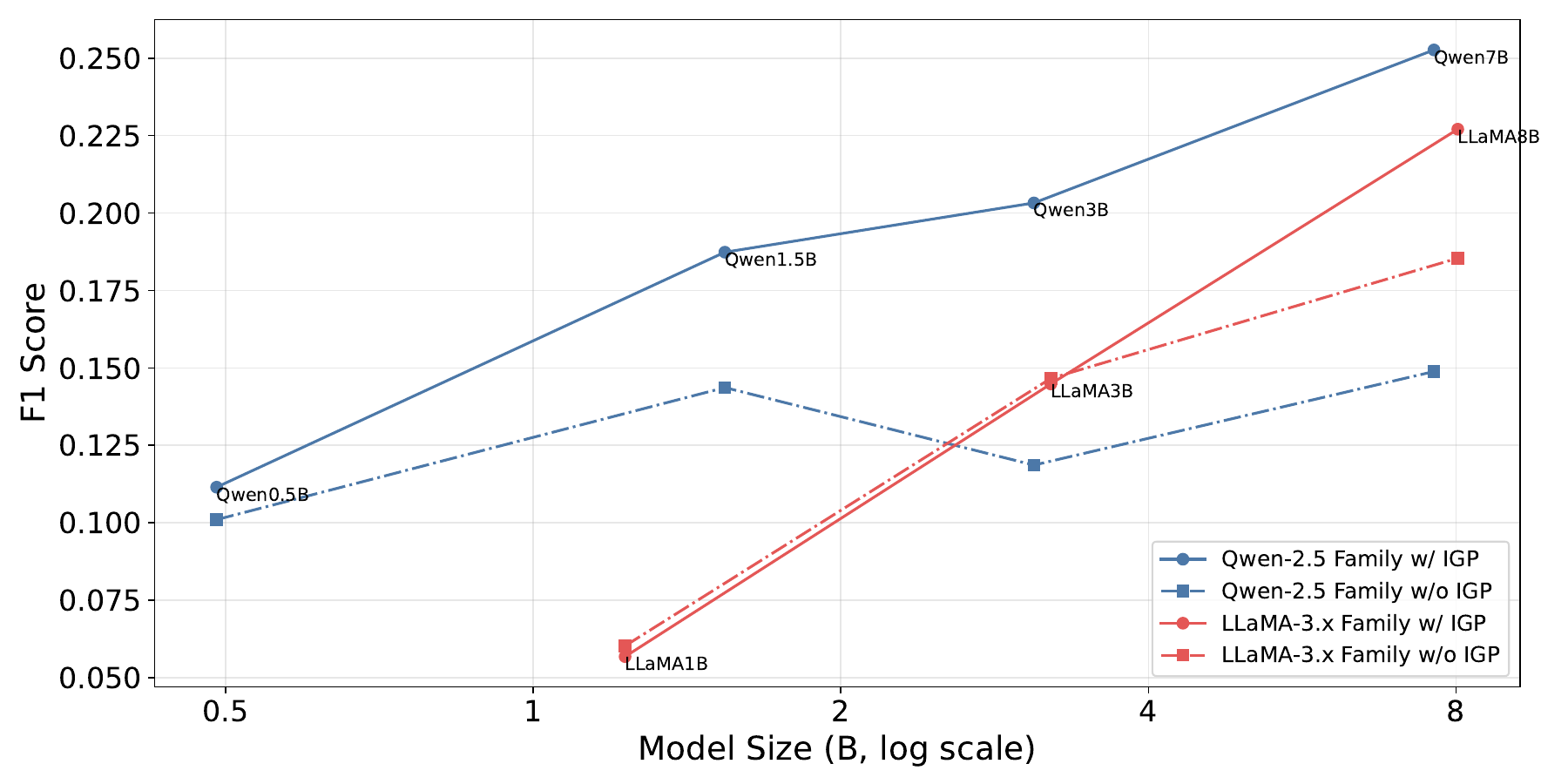}
\caption{\textbf{Effects of model family and scale with/without IGP under a tight evidence budget (\TopM$=1$) on NQ, with BM25 as the baseline retriever.}}
\label{fig:model-size-f1}
\end{figure}

Using the NQ dataset with BM25 as the first-stage retriever, we keep the corpus, candidate size, prompts, and decoding protocol fixed (\autoref{sec:exp_setup}) and vary only the generator family/scale.
We evaluate two instruction-tuned families (\textbf{Qwen2.5-Instruct} and \textbf{Llama-3.x-Instruct}) across multiple parameter scales, under a tight evidence budget \TopM$=1$.
For each generator, we compare standard RAG (\textbf{w/o IGP}) against RAG with Information Gain Pruning (\textbf{w/ IGP}) as a black-box rerank replacement (requiring only step-wise logits or \TopK log-probabilities; \autoref{sec:method_nu}--\autoref{sec:method_ig}).
Results are shown in \autoref{fig:model-size-f1}.

\paragraph{Small-model \,+\, IGP can outperform a larger model without IGP.}
A notable finding from \autoref{fig:model-size-f1} is that \textbf{Qwen1.5B with IGP (solid curve)} achieves higher end-to-end F1 than \textbf{Qwen7B without IGP (dashed curve)}.
This indicates that improving \emph{evidence utility} via IGP can compensate for (and in some cases exceed) gains from scaling the generator alone.
Practically, this suggests a strong efficiency lever: one can obtain \emph{large-model-level} (or better) performance using a smaller generator by upgrading the evidence-admission policy.

\paragraph{RAG + IGP exhibits a nearly log-linear scaling trend.}
When plotting F1 against model size on a log scale (x-axis in \autoref{fig:model-size-f1}), the \textbf{w/ IGP} curves rise in an approximately \textbf{log-linear} manner across scales for both families:
performance improves steadily as the generator grows, and the curve remains close to a straight-line trend in the log-size coordinate.
In contrast, the \textbf{w/o IGP} curves are consistently lower and can show weaker effective scaling because low-utility evidence under a fixed budget can introduce redundancy/conflicts that limit the generator from fully realizing its capacity.

\paragraph{Family robustness: consistent gains across Qwen2.5 and Llama-3.x.}
Across both model families, enabling IGP shifts the curve upward, demonstrating that the uncertainty-reduction-based utility signal generalizes beyond a single backbone.
Meanwhile, absolute performance differs by family at comparable scales (Qwen2.5 is generally stronger in our setting), indicating that generator choice sets the capability ceiling, while IGP provides a complementary, deployment-friendly improvement by filtering weak/conflicting evidence before context injection.

\subsection{Takeaways}
\label{sec:takeaways}
\begin{itemize}
  \item \textbf{RQ1 (Quality--cost trade-off):}
  IGP achieves a clear \emph{win--win}: it improves or maintains end-to-end F1 while reducing input-token cost (yielding a stronger Pareto frontier), whereas reranking-only brings limited gains.  

  \item \textbf{RQ2 (Mechanism and sensitivity):}
  Gains primarily come from \emph{admission control (pruning)} rather than reordering alone; performance shows a stable sweet spot for $T_p$ and plateaus for moderate $(K,MT)$, with $MT$ more influential for ranking stability (e.g., \autoref{sec:ablation_results}, \autoref{fig:ig-thres-f1}, \autoref{fig:topk-maxtk-combined}).

  \item \textbf{RQ3 (Relevance $\neq$ utility):}
  Offline relevance metrics (NDCG) are weakly correlated with (and can be negatively correlated with) end-to-end F1, especially under multi-evidence budgets, confirming that better relevance ranking does not guarantee better generation (e.g., \autoref{fig:retrieval-generation-corr}, \autoref{tab:ndcg-vs-f1}).

  \item \textbf{RQ4 (Retriever robustness):}
  IGP’s Pareto improvements persist when switching retrievers (BM25, sparse lexical term-matching $\rightarrow$ Contriever, dense bi-encoder embedding retrieval), indicating retriever-agnostic benefits from generator-aligned evidence utility (e.g., \autoref{tab:performance-contriever}, \autoref{fig:pareto-contriever}).

  \item \textbf{RQ5 (Generator robustness):}
  IGP is \textbf{model-agnostic} and \textbf{scales well}: it consistently improves end-to-end F1 across families and parameter scales, can allow a \textbf{smaller} generator (e.g., 1.5B) to \textbf{beat} a \textbf{larger} generator without IGP (e.g., 7B), and yields a \textbf{nearly log-linear} scaling behavior for RAG when evidence admission is aligned with generator utility(e.g., \autoref{fig:model-size-f1}).
\end{itemize}

\section{Conclusion}
\label{sec:conclusion}

Budgeted RAG is ultimately a decision problem: under a fixed context budget, the system must decide which evidence is worth paying for. In what follows, we summarize our main findings and contributions, highlight practical implications for deployment, and discuss limitations and future directions.

\subsection{Main Findings and Contributions}
We identify a persistent mismatch in budgeted RAG: \textbf{better retrieval relevance does not reliably translate into better end-to-end QA}, and the gap can widen when multiple passages are injected, where redundancy and mild conflicts become more common. This suggests that evidence selection should be optimized for \emph{utility to generation} rather than relevance alone.

To address this, we propose \textbf{Information Gain Pruning (IGP)}, a \textbf{generator-aligned} and \textbf{black-box friendly} drop-in replacement of the reranking stage. Instead of scoring passages only by relevance, IGP scores candidates by how much they \emph{increase the generator's decisiveness} under a fixed probing protocol, and prunes weak or harmful evidence before truncation.

Across multiple open-domain QA benchmarks, first-stage retrievers, and generator families/scales, IGP \textbf{improves or maintains answer quality while reducing final-stage input-token cost}, consistently strengthening the Pareto frontier of quality versus cost. In a representative multi-evidence setting, IGP improves average F1 by about \textbf{+12--20\%} while reducing final-stage input tokens by roughly \textbf{76--79\%} (relative to the retriever-only baseline). These gains indicate that \textbf{admission control (pruning)}---rather than reordering alone---is a dominant lever for improving the quality--cost trade-off in practical budgeted RAG.

\subsection{Practical Implications}
\label{sec:practical-implications}

IGP turns budgeted RAG into an explicit \emph{evidence admission-control} problem under a fixed context budget. It is easy to integrate because it \textbf{replaces only} \texttt{rerank} while keeping \texttt{truncate} and the original budget interface unchanged. This design fits common deployment workflows: IGP can be enabled as a policy swap in the reranking stage and evaluated through shadow runs or \textbf{A/B tests} without changing retrievers, prompts, or downstream executors. The pruning threshold $T_p$ provides a practical \textbf{quality--cost knob}: a higher $T_p$ yields more conservative admission (fewer passages admitted, lower context cost, typically higher stability), while a lower $T_p$ favors coverage when recall is prioritized.

\paragraph{When IGP is most effective.}
IGP is most effective when candidate pools contain \emph{semantically overlapping} or \emph{mildly conflicting} passages---a common failure mode in \emph{multi-evidence} RAG, where redundancy, ambiguity, and small inconsistencies can accumulate and destabilize generation. This pattern frequently arises in enterprise knowledge-base or policy QA, customer-support troubleshooting, and long-document QA over manuals/specs, where retrieved candidates are often highly similar paraphrases or differ in conditional scope and numeric details.

\paragraph{Operational control and fallback strategy.}
The system can implement lightweight admission-control fallbacks based on observable signals:
(i) \textbf{pass-rate gating}: monitor the fraction of candidates passing the threshold ($IG \ge T_p$); if the pass rate falls below a preset range (too few passages admitted), fall back to relevance-based reranking or disable pruning for that query while preserving the same \texttt{truncate} contract;
(ii) \textbf{uncertainty-triggered gating}: enable IGP primarily for queries with high baseline uncertainty (e.g., large $\widehat{NU}(q;\phi,K)$ under the probing prompt), where ambiguity and conflict sensitivity are more likely to matter.

\paragraph{Monitoring signals.}
Beyond acceptance statistics (e.g., pass rate under $T_p$), deployments can monitor \textbf{IG distribution drift} (e.g., mean/quantiles of $IG$ over time) to detect retriever/corpus changes or prompt/interface shifts that alter the scale or reliability of the utility signal. When drift or overly conservative admission is observed, the system can retune $T_p$ within a narrow range or temporarily fall back to relevance-based reranking, without changing the downstream budget executor.

\paragraph{Cost accounting in deployment.}
IGP adds selection-time probing, but this overhead is \emph{parallelizable} and can be amortized via batching/caching; critically, by pruning low-utility evidence it \textbf{reduces the final-stage context length}, which is the dominant factor for answer-generation cost under a fixed budget interface.

\subsection{Limitations and Future Work}
IGP uses uncertainty reduction as a generator-aligned proxy for evidence utility, which \textbf{does not guarantee correctness}. In particular, uncertainty can be reduced by \emph{misleading but confident} evidence, so IGP should be viewed as a \textbf{utility signal}, not a truth estimator. A practical direction is to \textbf{pair IGP with lightweight guardrails} such as source reliability constraints, consistency checks across admitted passages, or post-generation verification/answer checking when high stakes warrant it.

In addition, IGP estimates utility via \textbf{single-passage gain against a no-evidence baseline}, and thus does not explicitly model multi-evidence interactions (redundancy and complementarity). A natural extension is \textbf{set-aware selection}, e.g., computing \emph{conditional} information gain given an already selected set, or applying greedy/submodular-style selection with redundancy penalties (e.g., de-duplication) while preserving the deployment-friendly philosophy of modifying only the evidence-selection stage. These extensions may better capture joint evidence effects while retaining the modularity and black-box compatibility that make IGP practical.

\clearpage
\appendix

\section{Illustrative Example: Relevance $\neq$ Evidence Utility.}

\begin{figure}[pos=htbp]
\centering
\includegraphics[width=0.8\linewidth]{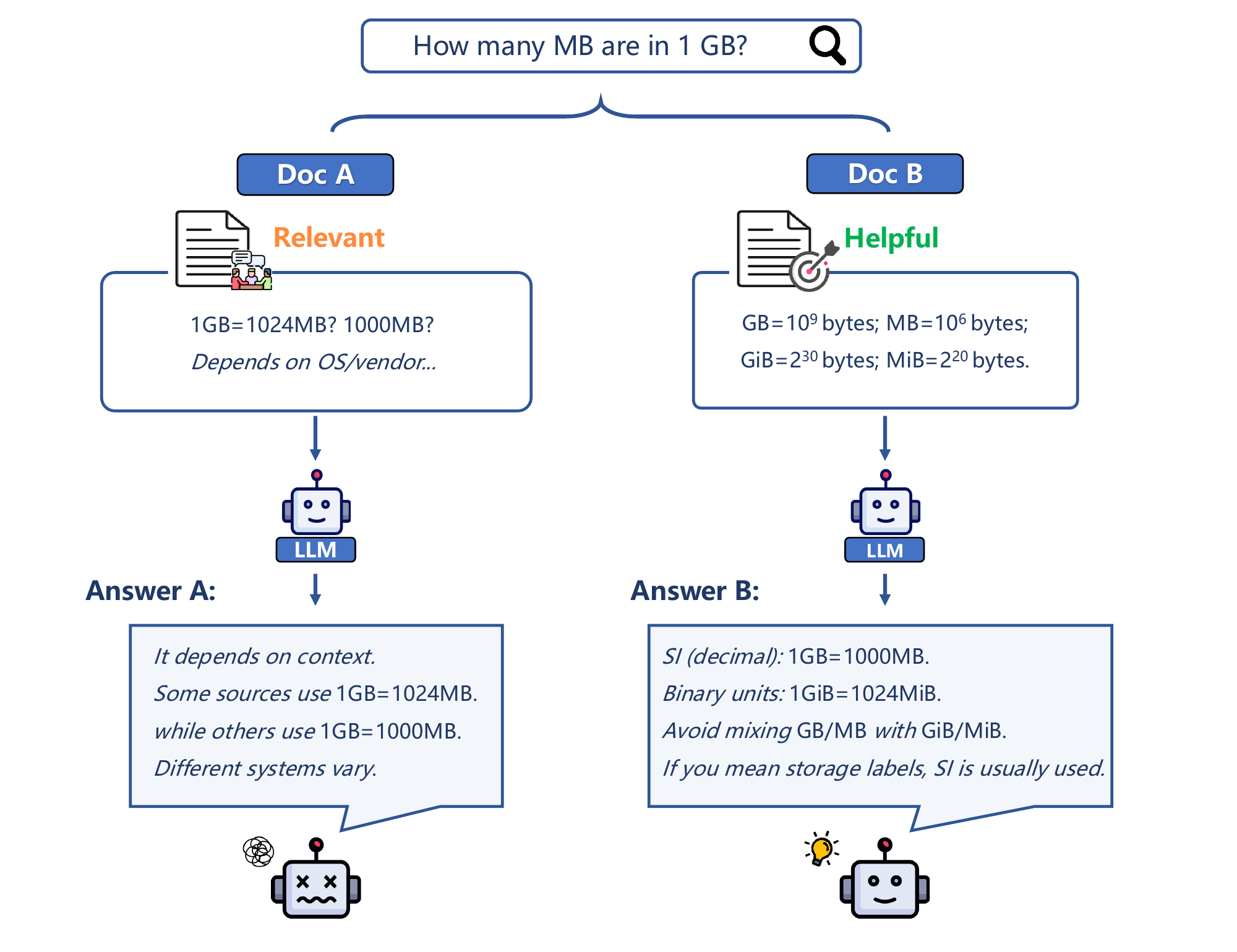}
\caption{\textbf{Illustrative Example: Relevance $\neq$ Evidence Utility.}
A relevant passage may introduce vague wording or parallel claims, increasing uncertainty in the generation distribution; in contrast, clearer and more decisive evidence often reduces generation uncertainty more effectively.}
\label{fig:relevance-vs-helpfulness}
\end{figure}

\clearpage
\bibliographystyle{apacite} 
\bibliography{reference}

\end{document}